\documentclass[11pt]{article}

\usepackage[letterpaper]{geometry}
\RequirePackage{amsthm,amsmath,amsfonts,amssymb}
\RequirePackage[authoryear]{natbib}%% uncomment this for author-year citations
\setcitestyle{authoryear,open={(},close={)}} 

\RequirePackage[colorlinks,citecolor=blue,urlcolor=blue]{hyperref}
\RequirePackage{graphicx}

\usepackage{amsmath,amsfonts,bm}

\def\eqref#1{equation~\ref{#1}}
\def\1{\bm{1}}

\DeclareMathAlphabet{\mathsfit}{\encodingdefault}{\sfdefault}{m}{sl}
\SetMathAlphabet{\mathsfit}{bold}{\encodingdefault}{\sfdefault}{bx}{n}

\def\namedlabel#1#2{\begingroup
    #2%
    \def\@currentlabel{#2}%
    \phantomsection\label{#1}\endgroup
}

\usepackage{amsmath, amssymb, amscd, amsthm, amsfonts,pbsi,aurical}
\usepackage{amssymb,amsthm,amsfonts,amsbsy,latexsym,amsxtra}
\usepackage{dsfont,todonotes}
\usepackage{graphicx,xcolor,color,bm,authblk}
\usepackage{mathtools,enumitem,todonotes,soul}
\usepackage[T1]{fontenc}
\usepackage[utf8]{inputenc}
\usepackage{float,mdframed}
\usepackage{mathrsfs} 
\usepackage{algorithm}
\usepackage{tikz}
\usepackage{float,mdframed}
\usepackage{algorithm}
\usepackage{algorithmic}

\usepackage{float,mdframed}
\usepackage{minitoc}

\numberwithin{equation}{section}

\makeatletter
\def\namedlabel#1#2{\begingroup
    #2%
    \def\@currentlabel{#2}%
    \phantomsection\label{#1}\endgroup
}
\makeatother

\newtheorem{Theorem}{Theorem}[section]
\newtheorem{Assumption}{Assumption}[section]
\newtheorem{Proposition}{Proposition}[section]

\newcommand{\mbb}{\mathbb}
\newcommand{\mbf}{\mathbf}

\newcommand{\mcal}{\mathcal}

\usepackage{wrapfig}
\usepackage{microtype}
\usepackage{graphicx}
\usepackage{enumitem}
\usepackage{booktabs}
\usepackage{enumitem}
\usepackage{tcolorbox}
\tcbuselibrary{skins,breakable}

\usepackage{float}
\usepackage{subfigure}
\usepackage{subcaption}
\usepackage{dsfont}

\allowdisplaybreaks

\begin{document}
\title{LOD: Latent Objective Discovery in Heterogeneous Multi-Agent Reinforcement Learning}

\author{Ke Sun}
\affil{University of Pennsylvania}

\author{Zhaoyang Shi\thanks{Corresponding author. Email: zyshi@fudan.edu.cn}}
\affil{Center for Applied Mathematics, Fudan University}

\maketitle

\begin{abstract}
In multi-agent reinforcement learning~(MARL) settings, agents are often heterogeneous with diverse intrinsic utility functions. In this study, we introduce \textbf{L}atent \textbf{O}bjective \textbf{D}iscovery~(\textbf{LOD}) as a framework for decomposing agents' distinct utilities into mixed underlying objectives in a low-dimensional latent space. Latent objective discovery aims to capture flexible and abstract coordination patterns among agents and each abstract objective is associated with a shared value function. During learning, all the learned values are aggregated by each agent according to their personalized weights learned in an adaptive manner. Therefore, agents inherit objective-level estimates for policy updates and value learning, enabling structured information sharing without requiring access to other agents’ policies. Algorithmically, we incorporate our approach into Multi-Agent Proximal Policy Optimization~(MAPPO) to exploit this structure. Theoretically, we establish convergence guarantees under linear function approximation within the actor-critic framework. Empirically, we extensively validate the advantage of introducing latent objective discovery in popular multi-agent testbeds with heterogeneous agents. 
\end{abstract}

\section{Introduction}

%%% background and motivation
Learning and understanding the \textit{heterogeneous behaviors} of agents in multi-agent reinforcement learning~(MARL) systems~\citep{busoniu2008comprehensive,zhang2021multi,albrecht2024multi} plays an increasingly crucial role in reliable deployment of reinforcement learning~(RL)~\citep{sutton1998reinforcement}. As a fundamental characteristic in MARL, heterogeneity~\citep{bennett2024heterogeneity,zhong2024heterogeneous} has recently been classified into five types: observation heterogeneity, response transition heterogeneity, effect transition heterogeneity, objective heterogeneity, and policy heterogeneity~\citep{hu2025heterogeneity}, where objective heterogeneity is associated with \textit{agents' reward functions} and is the focus in this study. In many MARL applications, such as sensor networks~\citep{rabbat2004distributed}, smart grid~\citep{dall2013distributed,zhang2018dynamic}, intelligent transportation systems~\citep{adler2002cooperative,zhang2018dynamic2}, robotics~\citep{corke2005networked}, economics~\citep{mazumdar2025tractable,tesauro2002pricing,ruadulescu2020multi}, traffic signal control~\citep{liu2023gplight}, and large language model fine-tuning~\citep{ma2024coevolving,liu2025uncertainty}, agents are not perfectly homogeneous and often optimize compatible yet only partially aligned objectives. Such objectives may be unobserved and arise from differences in attributes, expertise, or preferences among agents, leading to intrinsically distinct or structured utilities, giving rise to objective heterogeneity in complex MARL systems.

\noindent \textbf{Limitations of Existing MARL Paradigms for Objective Heterogeneity.} Despite the increasing importance of accounting for heterogeneity, objective heterogeneity remains a fundamental yet underexplored challenge in existing MARL paradigms. As a canonical paradigm, centralized training with decentralized execution~(CTDE)~\citep{amato2024introduction,oliehoek2008optimal,rashid2020monotonic,foerster2018counterfactual,mahajan2019maven,du2019liir} trains centralized critics with access to global information, while allowing the agent to act independently. Nevertheless, CTDE methods are predominantly designed for fully cooperative tasks with \textit{a shared global reward function} to simplify the analysis~\citep{panait2005cooperative,tuyls2012multiagent,zhang2021multi,amato2024introduction}, and thus do not explicitly account for agents with distinct underlying objectives or preferences. Consequently, homogeneous MARL algorithms could have restricted performance and applicability~\citep{zhong2024heterogeneous}, and also pose critical challenges for multi-agent credit assignment~\citep{chang2003all,foerster2018counterfactual}, i.e., identifying individual agents' contribution to the team's success. Moreover, fully decentralized MARL with networked agents~\citep{varshavskaya2009efficient,zhang2018fully,zhang2021multi} constitutes a complementary paradigm based on local centralization, accommodating heterogeneous rewards among agents under graph-structured interactions. However, the network topology deployed in this MARL paradigm is either explicitly pre-specified or heuristically defined by a local neighborhood among agents. We further provide an extended discussion of related work in Appendix~\ref{app:relatedwork}. These limitations expose a fundamental challenge in MARL: 
% how to adaptively learn an objective-aware coordination structure that reflects agents' heterogeneous utilities, which leaves a fundamental question unresolved:
\begin{center}
% \textit{How can we develop a principled paradigm to uncover objective-aware coordination structures among heterogeneous agents?
% }
\textit{How can we uncover objective-aware coordination structures among heterogeneous agents?}
\end{center}
%%% what we have done and performance
\textbf{Our Contribution: Latent Objective Discovery~(LOD) via Mixed Centralization.} To characterize the objective heterogeneity among agents and the structure of their effective coordination, we aim to infer hidden objectives in a low-dimensional latent reward space, where each discovered objective represents a different reward-aligned behavioral abstraction, adaptive to agents' heterogeneity. The latent objectives are shared among agents with different connecting weights, adaptive to their distinct attributes, preferences, and utilities. This strategy of leveraging mixing weights is in the spirit of the widely adopted \textit{mixed membership} concept in modeling heterogeneous generative densities \citep{fan2026optimal,shi2026population}, the underlying structure~\citep{erosheva2004mixed,airoldi2014introduction}, and the social network~\citep{airoldi2008mixed,graham2020network,jin2024mixed}. We realize latent objective discovery through \textit{mixed centralization}, which can be viewed as a strict generalization of CTDE. Our LOD can also be viewed from the perspective of \textit{networked agents}, where agents coordinate in a time-evolving network and heterogeneity of agents is naturally encoded in different clusters of the network. Unlike the classical networked MARL~\citep{zhang2018fully}, the network topology induced in our paradigm is latent, non-local, task-adaptive and statistically principled. By enabling each agent to share multiple objectives, our method is capable of capturing more complicated underlying coordination patterns, contributing to role differentiation, modularity, and generalization in real-world multi-agent systems. Algorithmically,  we introduce latent objective discovery into the state-of-the-art MARL algorithms, such as MAPPO. Our contributions can be summarized as follows:
% The contributions of our study can be succinctly summarized as follows:
\begin{itemize}[itemindent=\parindent,leftmargin=*, topsep=0pt, partopsep=0pt,itemsep=2pt,parsep=2pt]
    \item We introduce latent objective discovery into MARL with objective heterogeneity via a mixed centralization, adaptively capturing non-local and latent coordination patterns.
    
    \item We propose a practical actor-critic algorithm that effectively infers latent and overlapping objectives in MARL, with favorable theoretical properties.
    
    \item We empirically verify the benefits of latent objective discovery in MARL across various environments where agents inherently have distinct goals and incentives.
\end{itemize}

% \vspace{-7mm}
\section{Our Method: Latent Objective Discovery~(LOD) in MARL}\label{sec:LOD}

% \label{sec:preliminary_MDP}

\noindent \textbf{Preliminaries.} We study a Markov game~\citep{littman1994markov}, also known as a stochastic game~\citep{shapley1953stochastic}, which is defined by a tuple $\langle \mcal{N}, \mathcal{S}, \{\mathcal{A}^i\}_{i \in \mcal{N}}, P, \{R^i\}_{i \in \mcal{N}}, \gamma \rangle$ with agents $i \in \mcal{N} = \{1, ..., N\}$, the system state space $\mcal{S}$, the joint action space $\mcal{A} = \mcal{A}^1 \times \ldots \times \mcal{A}^N$, and the discount factor $\gamma$. $P:\mcal{S}\times \mcal{A} \rightarrow \Delta(\mcal{S})$ is the system state transition kernel and $R^i: \mcal{S}\times \mcal{A} \rightarrow \mbb{R}$ is the local (or shared) reward function for the agent $i$, giving the individual reward $r_t^i$ at the time step $t$. The individual policy is denoted by $\pi^i: \mcal{S} \rightarrow \Delta(\mcal{A}^i)$ and the joint policy $\boldsymbol{\pi}$ is defined by $\boldsymbol{\pi}(\cdot|s_t) = \prod_{i=1}^N \pi^i(\cdot|s_t)$. We thus define the average and individual state-action value functions: $Q^{\boldsymbol{\pi}}_{\text{ave}}(s, \boldsymbol{a}) = \mbb{E}_{\boldsymbol{\pi}}\left[\frac{1}{N}\sum_{i=1}^N \sum_{t=0}^\infty \gamma^t r_t^i | s_0=s, \boldsymbol{a}_0=\boldsymbol{a}\right]$ and $Q^{\boldsymbol{\pi}}_i(s, \boldsymbol{a}) = \mbb{E}_{\boldsymbol{\pi}}\left[\sum_{t=0}^\infty \gamma^t r_t^i | s_0=s, \boldsymbol{a}_0=\boldsymbol{a}\right]$ for $i \in \mathcal{N}$ given the system state $s$ and the joint action $\boldsymbol{a}=(a^1, \ldots, a^N)$. Notably, we allow each agent to have its own  reward function to accommodate the objective heterogeneity, which breaks the shared reward assumption and extends the cooperative settings commonly used in Dec-POMDP~\citep{oliehoek2008optimal,oliehoek2016concise}. The network topology is not pre-specified as adopted in networked MARL~\citep{zhang2018fully} and we will construct it based on the agents' reward signals to discover their latent objectives. We assume full observability in the theoretical derivations for tractable results, but we compensate for partial observability in implementations by employing RNN that is commonly adopted by existing MARL work, e.g., QMIX~\citep{rashid2020monotonic} and MAPPO~\citep{yu2022surprising}.

% \textbf{Heterogenity.} Rather than treating rewards as scalar signals, we view reward functions as elements in a function space and aim to learn low-dimensional representations that capture shared structure across agents. This perspective enables us to uncover latent objectives underlying heterogeneous agent behaviors. 

% We use the term latent goals to refer to implicit, reward-induced coordination objectives that emerge from agent interactions, rather than explicitly specified or symbolic task goals.

\subsection{Framework: Mixed Centralization for Latent Objectives}\label{sec:method_mixedcentral}

\textbf{Mixed Centralization and Its Motivation.} In the canonical CTDE paradigm, centralized critic methods, e.g., \cite{lowe2017multi,foerster2018counterfactual,yu2022surprising}, assume a single shared critic function among agents and are seen as a hard global centralization. Such a globally shared critic, however, does not explicitly capture the natural heterogeneity in agents’ underlying objectives, roles, and preferences. In contrast, we propose a mixed centralization regime that rests on multiple centralized critics, where each agent has its weights to \textit{all} critics rather than being tied to a single centralized critic. The mixed centralization regime is rooted in a  widely adopted  concept called \textit{mixed membership}. For example, according to people's purchase behaviors, they do not strictly belong to just one customer type. Instead, they are connected to core features of multiple consumer groups with distinct weights, such as tech enthusiasts, eco-conscious buyers, and budget-conscious shoppers.

\begin{wrapfigure}{r}{0.55\textwidth}
% \begin{figure}[htbp]
\centering
\vspace{-0.2in}
\includegraphics[width=0.55\textwidth,trim=80 50 30 0,clip]{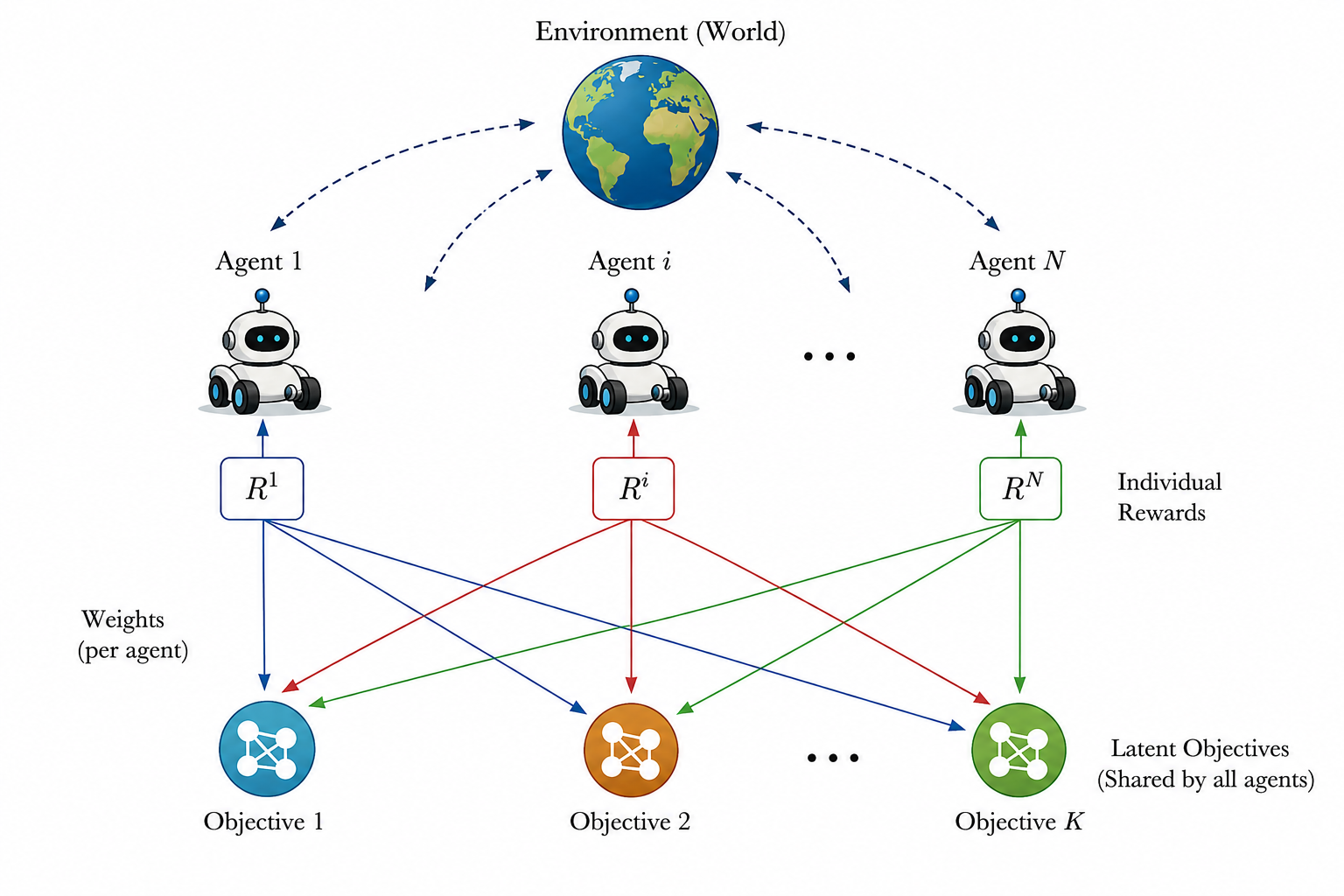}
\vspace{-0.2in}
\caption{Illustration of Latent Objective Discovery.}
\vspace{-0.2in}
\label{fig:illustration}
% \end{figure}
\end{wrapfigure}
\textbf{Multiple Latent Objectives in MARL.} The primary goal of mixed centralization is to discover multiple latent objectives in MARL, where agents typically possess different utility functions measured by distinct reward functions $\{R^i\}_{i\in\mcal{N}}$. Mathematically, assume that there are $K$ latent objectives among $N$ agents and denote $\gamma_i(k) \geq 0$ as the weight for the agent $i$ to the $k$-th latent objective, with $\sum_{k=1}^K\gamma_i(k)=1$ for $\forall i \in \mcal{N}$. The latent multiple objective paradigm in MARL is illustrated in Figure~\ref{fig:illustration}. Intuitively, the discrepancy among the rewards that agents receive reflects their underlying incentives and objectives in their coordination to achieve the pre-specified team goal. Here are some examples:
\begin{enumerate}[label=\textcircled{\arabic*},itemindent=\parindent,leftmargin=*, topsep=0pt, partopsep=0pt,itemsep=2pt,parsep=2pt]
    \item In cooperative navigation tasks, agents are required to cover areas 1 and 2~(objective 1) while avoiding collisions~(objective 2). Thus, the global reward can be decomposed into these two objectives, where each agent's reward is viewed as a weighted sum to positively contribute to both objectives with the weights determined by their spatial positions and different attributes.
    
    \item In cooperative and competitive games with heterogeneous agents, incentives of each agent's behaviors can be grouped into three categories: maximizing the individual rewards, cooperating with allies, and suppressing opponents’ advantages. These objectives are shared across agents but contribute differently to each agent due to distinct preferences, skills, and expertise.

    \item In agentic RL for task-solving, LLM agents are often assigned specialized roles such as planner, executor, critic, verifier, or tool-user. These role-specific rewards reflect heterogeneous incentives while jointly contributing to the global task objective.

    % \item In StarCraft tasks, despite the presence of a global reward and a shared team goal to accomplish, effective coordination relies on multiple latent objectives, including maximizing damage output, maintaining formation and positioning, and ensuring unit survivability. These objectives are shared across all agents but they exhibit different degrees of participation depending on their attributes and battlefield context. 
\end{enumerate}

\textbf{Mixed Centralization for Multiple Latent Objectives.} We provide an algorithmic perspective to achieve latent objective discovery in the general actor-critic framework via mixed centralization. Denote state-action value functions $\{Q^{\boldsymbol{\pi}}_i\}_{i \in \mcal{N}}$ and parameterized policies $\{\pi_{\theta_i}\}_{i \in \mcal{N}}$. By the policy gradient theorem~\citep{sutton1999policy}, the policy gradient of each agent should depend on each critic $Q^{\boldsymbol{\pi}}_i$ to maximize their own objective $\mathcal{J}(\theta_i)$. The standard centralized critic method in CTDE replaces $Q^{\boldsymbol{\pi}}_i$ by a centralized (averaged) critic $Q^{\boldsymbol{\pi}}_\omega$ parameterized by $\omega$, yielding the gradient of each $\pi_{\theta_i}$:
\begin{align}\label{eq:gradient_critic}
    \nabla_{\theta_i} \mathcal{J}\left(\theta_i\right)=\mathbb{E}\left[\nabla_{\theta_i} \log \pi_{\theta_i}\left(a^i | s\right) Q^{\boldsymbol{\pi}}_\omega\left(s, \boldsymbol{a}\right)\right].
\end{align}
By contrast, our mixed centralization regime in principle requires $K$ mixed centralized critics denoted by $\{Q^{\boldsymbol{\pi}}_{\omega_i}\}_{i=1}^K$ and their weighted sum forms the policy gradient:
\begin{align}\label{eq:gradient_critic_mixed}
    \nabla_{\theta_i} \mathcal{J}_{\text{LOD}}\left(\theta_i\right)=\mathbb{E}\left[\nabla_{\theta_i} \log \pi_{\theta_i}\left(a^i | s\right) \left( \sum_{k=1}^K \gamma_{i}(k) Q^{\boldsymbol{\pi}}_{\omega_k}\left(s, \boldsymbol{a}\right) \right)\right],
\end{align}
where we replace the centralized critic $Q^{\boldsymbol{\pi}}_{\omega}$ in Eq.~\ref{eq:gradient_critic} by the weighted form $\sum_{k=1}^K \gamma_{i}(k) Q^{\boldsymbol{\pi}}_{\omega_k}$ to perform mixed centralization. To avoid degeneration, each $Q^{\boldsymbol{\pi}}_{\omega_k}$ should reflect distinct objectives shared among agents, which is specifically achieved by decomposing the reward signals in the latent space. 

\subsection{Reward Decomposition for Latent Objective Discovery}\label{sec:method_LOD}

\noindent \textbf{Latent Reward Decomposition.} Let $\{\boldsymbol{R}_t\}_{t>0}$ be the $N$-dimensional time series with the random vector $\boldsymbol{R}_t = [R_t^1, \ldots, R_t^N]^\top \in \mbb{R}^N$, where $R_t^i := R^i(s_t, \boldsymbol{a}_t)$ is the stochastic reward for the agent $i$ at the time step $t$. For an integer $K < N$, let $\gamma_i = [\gamma_{i}(1), \ldots, \gamma_i(K)]^\top \in \mbb{R}^K$ be the weight vector for the agent $i$, forming a loading weight matrix $\Gamma = [\gamma_1, \ldots, \gamma_N]^\top \in \mbb{R}^{N \times K}$. Let $\{\boldsymbol{F}_t\}_{t>0}$ be a $K$-dimensional covariance-stationary time series  with $\boldsymbol{F}_t = [F_t^1, \ldots, F_t^K]\in \mbb{R}^K$. Assume a population rank-$K$ factor model~\citep{lawley1962factor,harman1976modern,kline2014easy} on the reward time series $\{\boldsymbol{R}_t\}_{t>0}$ and there exists a \textit{time-homogeneous} weight matrix $\Gamma$ such that\
\begin{align}\label{eq:reward_decomposition}
    \boldsymbol{R}_t = \Gamma \boldsymbol{F}_t + \boldsymbol{\epsilon}_t,
\end{align}
where $\{\boldsymbol{\epsilon}_t\}_{t>0}$ is the white noises.
% with $\mbb{E}[\boldsymbol{\epsilon}_t]=\boldsymbol{0}_N$, $\text{Cov}(\boldsymbol{\epsilon}_t, \boldsymbol{\epsilon}_{t'})=\sigma^2 I_N$ when $t'=t$, otherwise $\boldsymbol{0}_{N \times N}$. 
$\{\boldsymbol{F}_t\}_{t>0}$ are the latent $K$-dimensional factors and each latent time series $\{F_t^k\}_{t>0}$ for $k \in [K]$ corresponds to a distinct optimization signal shared across agents throughout the learning process, which we call a \textit{latent objective}. Under the linear factor model assumption in Eq.~\ref{eq:reward_decomposition}, each agent's reward time series can be expressed as a linear combination of $K$ latent objectives that jointly explain cross-agent reward variation over time. 

% All rewards are finally received by agents in a way of being aggregated to maximize the global average rewards. 

\noindent \textbf{Extension to Latent Critic Decomposition.} Define the latent objective critic $\boldsymbol{Q}_{\boldsymbol{F}}^{\boldsymbol{\pi}}(s, \boldsymbol{a}):=\mathbb{E}_{\boldsymbol{\pi}}\left[\sum_{t=0}^{\infty} \gamma^t \boldsymbol{F}_{t} \mid s_0=s, a_0=\boldsymbol{a}\right] \in \mathbb{R}^K$ and denote $\boldsymbol{Q}^{\boldsymbol{\pi}}(s, \boldsymbol{a}) = [Q_1^\pi(s, \boldsymbol{a}), \ldots, Q_N^\pi(s, \boldsymbol{a})]^\top \in \mathbb{R}^N$. Proposition~\ref{prop:critic_decomposition} shows that latent reward decomposition in Eq.~\ref{eq:reward_decomposition} implies a latent critic decomposition with the transitivity of the $\Gamma$-transformation from the reward space to the state-action value function space. The proof is provided in Appendix~\ref{app:proof_prop_criticdecompose}.
\begin{Proposition}[Latent Critic Decomposition]\label{prop:critic_decomposition}
    By latent reward decomposition in Eq.~\ref{eq:reward_decomposition}, we have
    \begin{align}\label{eq:critic_decomposition}
        \boldsymbol{Q}^\pi(s, \boldsymbol{a}) = \Gamma  \boldsymbol{Q}_{\boldsymbol{F}}^{\boldsymbol{\pi}}(s, \boldsymbol{a}).
    \end{align}
\end{Proposition}
Next, by applying the function approximation, the $k$-th element of $\boldsymbol{Q}_{\boldsymbol{F}}^{\boldsymbol{\pi}}$ is parameterized by $Q_{\omega_k}$. Proposition~\ref{prop:critic_decomposition} naturally leads to the mixed centralization of policy gradient in Eq.~\ref{eq:gradient_critic_mixed} of Section~\ref{sec:method_mixedcentral}: 
\begin{align}
    Q^{\boldsymbol{\pi}}_i(s, \boldsymbol{a}) \approx \sum_{k=1} \gamma_{i}(k) Q_{\omega_k}(s, \boldsymbol{a}).
\end{align}
\textbf{Online Estimation of $\Gamma$ and $\boldsymbol{F}_t$.} Next, we perform an online estimation of $\Gamma$ and $\boldsymbol{F}_t$ against the non-stationarity due to the constantly updated RL policies. Firstly, we perform rollout to collect the trajectory $\{s_t, \boldsymbol{a}_t, \boldsymbol{r}_t, s_{t+1}\}_{t=1}^T$ with the length $T$, where $\boldsymbol{r}_t = [r_t^1, \ldots, r_t^N]^\top \in \mbb{R}^N$  is the $N$-dimensional observed reward vector. We denote $\boldsymbol{r}_{1:T} = [\boldsymbol{r}_1, \ldots, \boldsymbol{r}_T]\in \mbb{R}^{N \times T}$ as the observed reward matrix and the latent objective matrix $\boldsymbol{f} = [\boldsymbol{f}_1, \ldots, \boldsymbol{f}_T] \in \mbb{R}^{K \times T}$ with $\boldsymbol{f}_t=[f_t^1, \ldots, f_t^K]^\top \in \mbb{R}^K$.  Estimating the loading matrix $\Gamma$ and the latent objective $\boldsymbol{f}$ reduces to a rank-$K$ low-rank approximation problem with convex constraints. The optimization is expressed as:
\begin{align}\label{eq:LOD_estimation}
    \min _{\Gamma, \boldsymbol{f}}\|\boldsymbol{r}_{1:T} -\Gamma \boldsymbol{f}\|_F^2 + \lambda \| \boldsymbol{f} \|_F^2 \quad \text { s.t. } \Gamma \geq 0, \Gamma \mathbf{1}=\mathbf{1}.
\end{align}
The solution is reminiscent of the truncated SVD and Principal Component Analysis~(PCA). In practice, we can perform alternating update.  By fixing $\boldsymbol{f}$, updating $\Gamma$ is equivalent to solving simplex-constrained least squares $\min_{\Gamma \in \Delta^{K-1}}  \|\boldsymbol{r}_{1:T} -\Gamma \boldsymbol{f}\|_F^2$. By fixing $\Gamma$, updating $\boldsymbol{f}$ reduces to a ridge-regularized least-squares estimation with the closed-form solution: $\boldsymbol{f} = (\Gamma^\top \Gamma + \lambda \mathbf{I})^{-1} \Gamma^\top \boldsymbol{r}_{1:T}$. The estimation of $\Gamma$ and $\boldsymbol{F}_t$ can be performed 
% either offline, using a large pre-collected dataset of length $T$ and holding the resulting estimates fixed during subsequent training, or 
online with $\Gamma$ updated according to $\Gamma_{t+1} = (1-\tau)\Gamma_t + \tau \Gamma$, where $\Gamma$ is the newly estimated matrix via Eq.~\ref{eq:LOD_estimation} and $\tau \in [0,1]$ controls the update rate. This estimation procedure is a key component of the LOD module and supports its integration into existing MARL algorithms, e.g., MAPPO described in Section~\ref{sec:algorithm}.

% The proposed objective is biconvex but not jointly convex. As such, the alternating minimization algorithm is not guaranteed to find the unique global optimum in general. Nevertheless, under standard block coordinate descent conditions, the objective value decreases monotonically, and provably converges to a stationary point~\citep{bai2003inferential,tseng2001convergence}.

\subsection{Algorithm}\label{sec:algorithm}

LOD is a general, modularized component that can be integrated into a wide range of MARL algorithms. In this study, we use MAPPO~\citep{yu2022surprising} as our primary instantiation, given its widespread adoption and strong empirical performance across diverse benchmarks~\citep{papoudakis2021Benchmarking}. We then discuss how LOD can be incorporated into a broader class of actor–critic algorithms.

\noindent \textbf{LOD-MAPPO Algorithm.} We parameterize $K$ state-value functions $\{V_{\omega_k}\}_{k=1}^K$ instead of $\{Q_{\omega_k}\}_{k=1}^K$ to facilitate Generalized Advantage Estimation~(GAE)~\citep{schulman2015high}. We also denote $o_t^i$ as the observation for agent $i$ at the time $t$ to accommodate partial observability. After performing the online estimation in Eq.~\ref{eq:LOD_estimation}, we discover the latent objective signal $f_t^k$ in the $k$-th objective for each step $t$ and then we can define the Temporal-Difference~(TD) error $\delta_t^k$ for $V_{\omega_k}$: $\delta_t^k=f_t^k+\gamma V_{\phi_k}\left(s_{t+1}\right)-V_{\phi_k}\left(s_t\right)$. Consequently, the $k$-th latent advantage estimate is $\hat{A}_t^k=\sum_{\ell=0}^{T-t}(\gamma \lambda)^{\ell} \delta_{t+\ell}^k$ via GAE for $k \in [K]$. For the agent $i$, we denote the personalized advantage estimate by $\bar{A}^i_t$, which is the linear combination between $\hat{A}_t^k$ and $\gamma_i$: $\bar{A}^i_t = \sum_{k=1}^K \gamma_i(k) \hat{A}_t^k$. The algorithmic description of LOD-MAPPO is provided in Algorithm~\ref{alg:LOD_MAPPO_short}. To improve memory and computational efficiency, we employ a multi-head with a shared backbone $h_\psi(s_t)$, with $V_{\omega_k}(s_t)=g_{\omega_k}\left(h_\psi(s_t)\right)$ for $k \in [K]$.
\begin{algorithm}[t!]
    \centering
	\caption{MAPPO Algorithm with Latent Objective Discovery~(on-policy)}
	\begin{algorithmic}[1] %[1] enables line numbers
    \STATE \textbf{Input}: Episodes $L$, batch size $B$, GAE parameter $\lambda$, steps per episode $T$, number of epochs $M$
		\STATE \textbf{Initialize}: policy networks $\{\pi_{\theta_i^0}\}_{i \in \mcal{N}}$, value networks $\{V_{\phi_k^0}\}_{k=1}^K$, rollout buffer $\mathcal{B}$
\FOR{$l = 1,2,\dots, L$}
    \STATE Reset the environment and obtain the initial state $s_1$ and observations $\{o_1^i\}_{i\in\mathcal N}$.
    \STATE Collect a set of trajectories by running the joint policy $\boldsymbol{\pi}_{\theta^{l-1}}=(\pi_{\theta_1^{l-1}}, \ldots, \pi_{\theta_N^{l-1}})$
     \STATE Push transitions $\{s_t, \{o_t^i\}_{i \in \mcal{N}}, \{a_t^i\}_{i \in \mcal{N}}, \{r_t^i\}_{i\in \mcal{N}}, s_{t+1}, \{o_{t+1}^i\}_{i \in \mcal{N}}\}_{t=1}^T$ into $\mathcal{B}$
     \STATE Update $\Gamma$ and decompose $\boldsymbol{f}$ by solving Eq.~\ref{eq:LOD_estimation} based on $\boldsymbol{r}_{1:T}$. {\color{gray} \#  Latent Objective Discovery.}
     \STATE  {\color{gray} \#  Mixed Centralized Advantage Function via GAE.}
      \FOR{$k = 1, \ldots K$}
        % \FOR{$t = 1,\ldots,T$}
        \STATE $\hat{A}_t^k=\sum_{\ell=0}^{T-t}(\gamma \lambda)^{\ell} \delta_{t+\ell}^k$, and $\delta_t^k=f_t^k+\gamma V_{\phi_k^{l-1}}\left(s_{t+1}\right)-V_{\phi_k^{l-1}}\left(s_t\right)$.
    \ENDFOR
    
   \FOR{$m=1, \ldots, M$}
   
   \STATE Sample a random minibatch of $B$ transitions from $\mathcal{B}$

    \STATE {\color{gray} \#  Actor update.}
    \FOR{$i \in \mcal{N}$}
        \STATE Update each policy by maximizing the PPO-Clip objective with respect to $\theta_i$:
        \vspace{-0.3cm}
    \begin{align*}
   \pi_{\theta_i^l} \leftarrow \arg\max_{\pi_{\theta_i}} \frac{1}{T} \sum_{t=0}^T
    \min\left(
  \rho_{t}(\theta_i) \bar{A}^i_t,\;
    \text{clip}(\rho_{t}(\theta_i),1-\epsilon,1+\epsilon)\bar{A}^i_t
    \right),
    \end{align*} 
    where $\rho_{t}(\theta_i)= \frac{\pi_{\theta_i}(a^i_t|o^i_t)}{\pi_{\theta_i^{l-1}}(a^i_t|o_t^i)}$ and $\bar{A}^i_t = \sum_{k=1}^K \gamma_i(k)\hat{A}^k_t$.  
    \ENDFOR
    \STATE {\color{gray} \#  Critic update.}
    \FOR{$k = 1, \ldots, K$}
    \STATE Compute the GAE return for the $k$-th objective by $\hat{R}_t^k = \hat{A}^k_t+ V_{\phi_k^{l-1}}(s_t)$
    \STATE Update the value function by $V_{\phi_k^l} \leftarrow \arg\min_{V_{\phi_k}}\frac{1}{T} \sum_{t=1}^T \left[(V_{\phi_k}(s_t)-\hat{R}^k_t)^2\right]$
    \ENDFOR
    \ENDFOR
    \STATE Clear the rollout buffer $\mathcal{B}$
    % \STATE $\Gamma_0 \leftarrow \Gamma_T$ and $\boldsymbol{f}_0 \leftarrow \boldsymbol{f}_T$ {\color{gray}\# Episode-level warm start}
\ENDFOR
	\end{algorithmic}
	\label{alg:LOD_MAPPO_short}
\end{algorithm}

\noindent \textbf{Broader Actor-Critic Algorithms.} Our latent objective discovery framework can accommodate a much broader actor-critic algorithm class, including both on-policy and off-policy branches. In Appendix~\ref{appendix:algorithm_generic}, we include a generic on-policy learning framework in Algorithm~\ref{alg:generic_on} that can also instantiate trust-region objective in TRPO~\citep{schulman2015trust}. We also include an off-policy learning framework in Algorithm~\ref{alg:generic_off}, where we reproject sampled rewards from the replay buffer to decompose them into latent objectives. The TD target $Q_k$ can be instantiated using either the target network in DDPG~\citep{lillicrap2015continuous} or clipped double Q-learning in TD3~\citep{fujimoto2018addressing}. We also present a generic actor-critic algorithm in the average reward setting in Algorithm~\ref{alg:generic_averagereward} of Appendix~\ref{appendix:algorithm_averagerewards}, which is the focus for the theoretical analysis in Section~\ref{sec:method_theory}.
% Notably, the proposed framework reduces to the corresponding vanilla actor–critic algorithms without latent objective discovery when $K=1$.  

\textbf{Extension to Shared Global Rewards via Agent-Specific Advantage Surrogates.} In many MARL applications, agents only have access to sparse and delayed rewards, such as a shared and trajectory-level reward when the episode completes, even though they may perform distinct roles and contribute differently to the final outcome. Without the heterogeneous rewards to perform LOD, we can still leverage the agent-specific local advantage function as a surrogate to measure the contribution of agent \(i\)'s action relative to its local alternatives, based on which we perform LOD method. This strategy is also commonly adopted by \cite{foerster2018counterfactual,zhang2018fully,zhong2024heterogeneous}, as the surrogate local reward signal. Particularly, the agent-specific advantage function is defined by
\begin{align}\label{eq:local_advantage}
A_i^{\boldsymbol{\pi}}(s, \boldsymbol{a})=Q^{\boldsymbol{\pi}}(s, \boldsymbol{a}) - \sum_{\tilde{a}^i\in\mcal{A}^i} \pi^i(\tilde{a}^i|s)Q^{\boldsymbol{\pi}}(s,(\tilde{a}^i,a^{-i})),
\end{align}
where $a^{-i}$ is the action vector of all agents except $a^i$. Inspired by \textit{difference rewards}~\citep{wolpert2001optimal}, this agent-specific advantage also has the same form of \textit{counterfactual baseline}~\citep{foerster2018counterfactual} and \textit{aristocrat utility}~\citep{wolpert2001optimal} to represent their heterogeneous incentives. We validate the efficacy of the local advantage surrogate in Section~\ref{sec:experiment_agenticRL} of our experiments.

\section{Theoretical Results}\label{sec:method_theory}
For theoretical analysis, we consider the average-reward setting~\citep{puterman2014markov,marbach2002simulation,abounadi2001learning,zhang2018fully,chen2023finite,wu2020finite}, where our policy optimization objective is $\max_{\theta}J(\theta)=\underset{T\rightarrow \infty}{\lim}~\frac{1}{T}\mbb{E}\left(\sum_{t=0}^{T-1}\frac{1}{N}\sum_{i\in\mcal{N}}r_{t+1}^i\right)$. Specifically, our theoretical results focus on the average-reward setting with a generic actor-critic algorithm, which we defer to Algorithm~\ref{alg:generic_averagereward} in Appendix~\ref{appendix:algorithms}. Under standard ergodicity assumptions, this average reward setting coincides with the normalized discounted reward setting in the limit as the discount factor $\gamma$ approaches one, and the corresponding policy gradients agree in this limit, providing theoretical justification for the discounted reward setting in our LOD. 

\begin{Assumption}\label{Assump_1}
    The latent factor $f_t^{k}$ and the instantaneous reward $r_t^i$ are uniformly bounded over all $1\le k\le K,1\le i\le N$ and $t\ge 0$.
\end{Assumption}
This assumption on the latent factor model is standard and mild. If considering finite state and action spaces, it is automatically satisfied.

\begin{Assumption}\label{Assump_2}
    The update of the policy parameter $\theta_t^i$ includes a local projection operator $\mcal{L}_0^i:\mbb{R}^{m_i}\rightarrow \Theta^i\subset\mbb{R}^{m_i}$, projecting any $\theta_t^i$ onto the compact set $\Theta^i$. Additionally, we assume $\Theta=\prod_{i=1}^{N}\Theta^i$ is large enough to include at least one local minimum of $J(\theta)$. 
\end{Assumption}
This projection assumption is commonly used and standard in many analyses for actor-critic algorithms in stabilizing the stochastic approximation. See \cite{zhang2018fully,bhatnagar2007naturalgradient,degris2012off,prasad2014actor}.

\begin{Assumption}\label{Assump_3}
    The stepsizes $\eta_{\omega,t}$, $\eta_{\theta,t}$ satisfy $\sum_t \eta_{\omega,t}=\sum_t \eta_{\theta,t}=\infty,\ \sum_t \eta_{\omega,t}^2+\eta_{\theta,t}^2<\infty$. Moreover, $\eta_{\theta,t}=o(\eta_{\omega,t})$.
\end{Assumption}
Here, the critic step updates faster than the actor step, which behaves like a two-time-scale algorithm as standard single-agent actor-critic algorithms.

\begin{Assumption}\label{Assump_4}
    The online weight estimator $\gamma_{it}$ is consistent in $t$, i.e., the weight vector $\gamma_{it}\rightarrow\gamma_i\in\mbb{R}^K$ for all $i\in\mcal{N}$.
\end{Assumption}
This assumption provides generality rather than restriction: our analysis accommodates any consistent estimator, including standard factor-based estimators under regularity conditions in Eq.~\ref{eq:LOD_estimation}.

\begin{Assumption}\label{Assump_5}
    The $Q$-function is parametrized as $Q_{\omega}(s, a)=\omega'\phi(s,a)$, where $\phi(s,a)=[\phi_1(s,a),\ldots,\phi_{m_{\Omega}}(s,a)]'$ is some feature vector associated with $(s,a)$. Moreover, $\phi(s,a)$ is uniformly bounded for any $(s,a)$. Define the feature matrix $\Phi\in\mbb{R}^{|\mcal{S}||\mcal{A}|\times m_{\Omega}}$ whose $m$-th column is equal to $[\phi_m(s,a),s\in\mcal{S},a\in\mcal{A}]'$ for $1\le m\le m_{\Omega}$. Then, $\Phi$ is assumed to have full column rank and $\Phi u\neq \mbf{1}$ for any $u\in\mbb{R}^{m_{\Omega}}$.
\end{Assumption}
Here, for the convenience of theoretical analysis, we only focus on the linear function approximation. Actually, under nonlinear function approximation, TD-learning-based policy evaluation may fail to converge~\citep{tsitsiklis1996analysis}. Based on such assumptions, most of the existing online actor-critic algorithms establish their theoretical convergence guarantees~\citep{konda1999actor,bhatnagar2007naturalgradient,bhatnagar2010actor,zhang2018fully}. As for the assumption on the feature matrix $\Phi$, it is also commonly adopted in many studies \cite{bhatnagar2007naturalgradient,tsitsiklis1996analysis,tsitsiklis1999average,zhang2018fully}.
% \iffalse
% \zy{to Ke, I know you mentioned this easier that it seems we are out of pages. If so, I will just write the theorems as our algorithm converges and write the detailed limits in the appendix then we don't need the following in the main paper} \ke{agree. try to keep the theory concise and avoid to introduce too many notations. If possible, focus on the convergence result and put all details in appendix.}
% \fi
 
\begin{Theorem}[Convergence of the critic step]\label{thm:convergen_omega}
    Under Assumptions \ref{Assump_1} and \ref{Assump_3} to \ref{Assump_5}, for any policy $\pi_{\theta}$ given, the factor parameter $\omega_{t}^{(k)}$ converges: $\lim_t\omega_{t}^{(k)}=\omega_{\theta}$ almost surely and the limit $\omega_{\theta}$ is the unique solution to
    \begin{align*}
        \Phi'D_{\theta}^{s,a}(T_{\theta}^{Q,(k)}(\Phi\omega_{\theta}^{(k)})-\Phi\omega_{\theta}^{(k)})=0,\ 1\le k\le K.
    \end{align*}
\end{Theorem}
For clarity, the full definitions of $D_{\theta}^{s,a}$, the Bellman operator
$T_{\theta}^{Q,(k)}$ are provided in
Appendix~\ref{app:theorydetails}.

\begin{Theorem}[Convergence of the actor step]\label{thm:convergence_theta}
    Under Assumptions \ref{Assump_1} to \ref{Assump_5}, the agent policy parameter $\theta_t^i$ converges almost surely to a point in the set of the asymptotically stable equilibria of
    \begin{align*}
        \dot{\theta}^i=\mcal{L}^i(\mbb{E}_{s_t\sim d_{\theta},a_t\sim \pi_{\theta}}(A_{t,\theta}^i\psi_{t,\theta}^i)),\ i\in\mcal{N}.
    \end{align*}
\end{Theorem}
The definitions of $A_{t,\theta}^i$, $\psi_{t,\theta}^i$, and the projection
operator $\mathcal{L}^i(\cdot)$ are provided in
Appendix~\ref{app:theorydetails}.

\section{Experiment}\label{sec:experiment}

In this section, we evaluate and analyze the effect of latent objective discovery via LOD-MAPPO on an array of multi-agent benchmarks with heterogeneous rewards: Multi-Agent Particle Environment~(MPE)~\citep{lowe2017multi,mordatch2018emergence}, Level-Based Foraging~(LBF)~\citep{albrecht2015game,albrecht2019reasoning}, Multi-Robot Warehouse~(RWARE)~\citep{christianos2020shared}, Agentic
RL tasks in Agent Lightning~\citep{masry2022chartqa,he2026agent}, including ChartQA and MuSiQue. These benchmarks are diverse in task difficulty, agent number, action type,  observation and action space dimensionality, and cooperation strategy required, and hence provide a comprehensive assessment of the effectiveness, stability, robustness, and generality of our methods. The selected benchmarks are illustrated in Figure~\ref{fig:testbeds} and detailed descriptions of each benchmark are provided in Appendix~\ref{appendix:experiment}. Our theoretical derivations assume full observability, a common model choice in \citep{zhang2018fully,zhong2024heterogeneous}, but our experiments compensate for partial observability by conditioning on past observations as adopted in MAPPO~\citep{yu2022surprising}, QMIX~\citep{rashid2020monotonic}, and COMA~\citep{foerster2018counterfactual}.

\begin{figure}[htbp]
    \centering
    \includegraphics[width=0.15\linewidth]{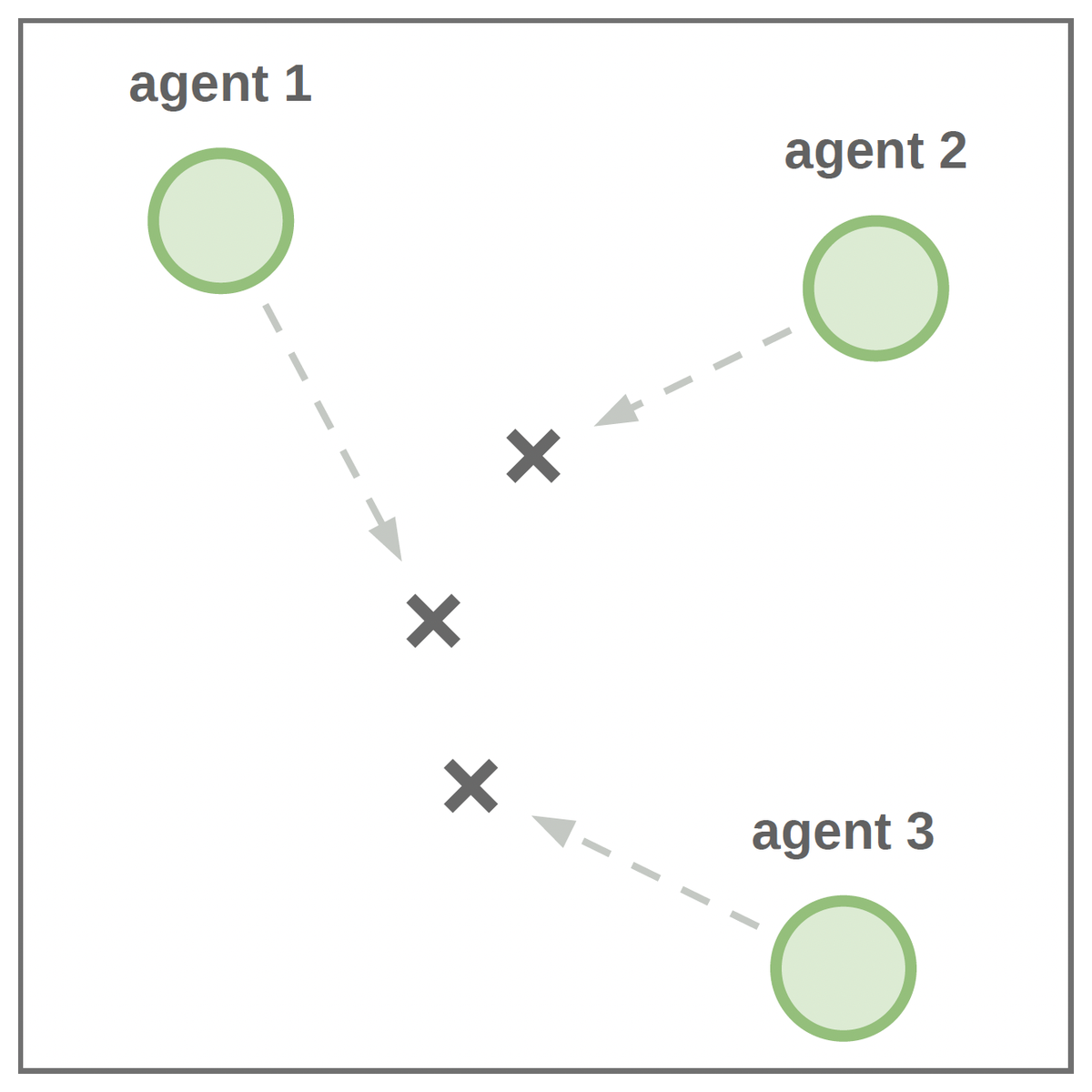}\hfill
    \includegraphics[width=0.15\linewidth]{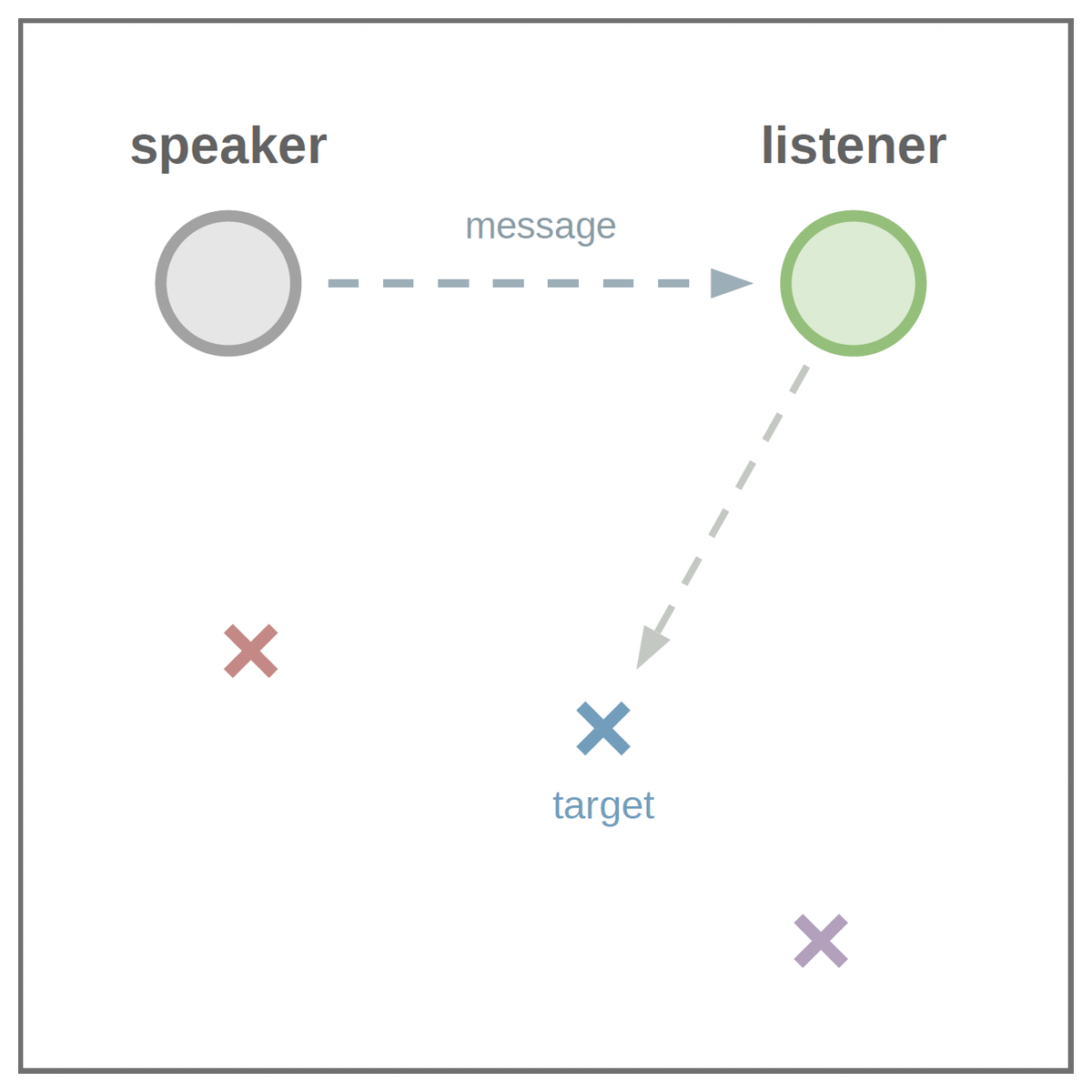}\hfill
    \includegraphics[width=0.15\linewidth]{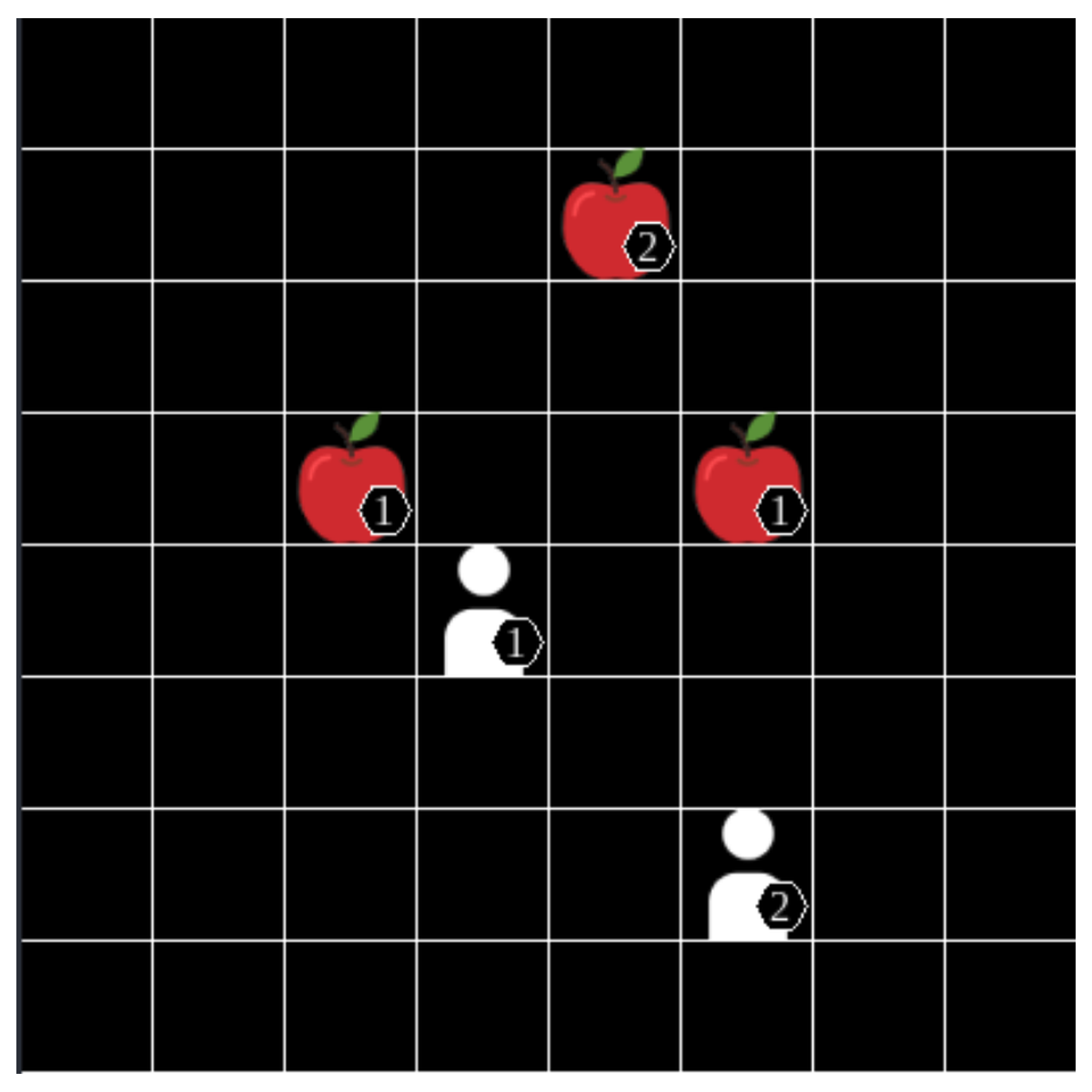}\hfill
    \includegraphics[width=0.15\linewidth]{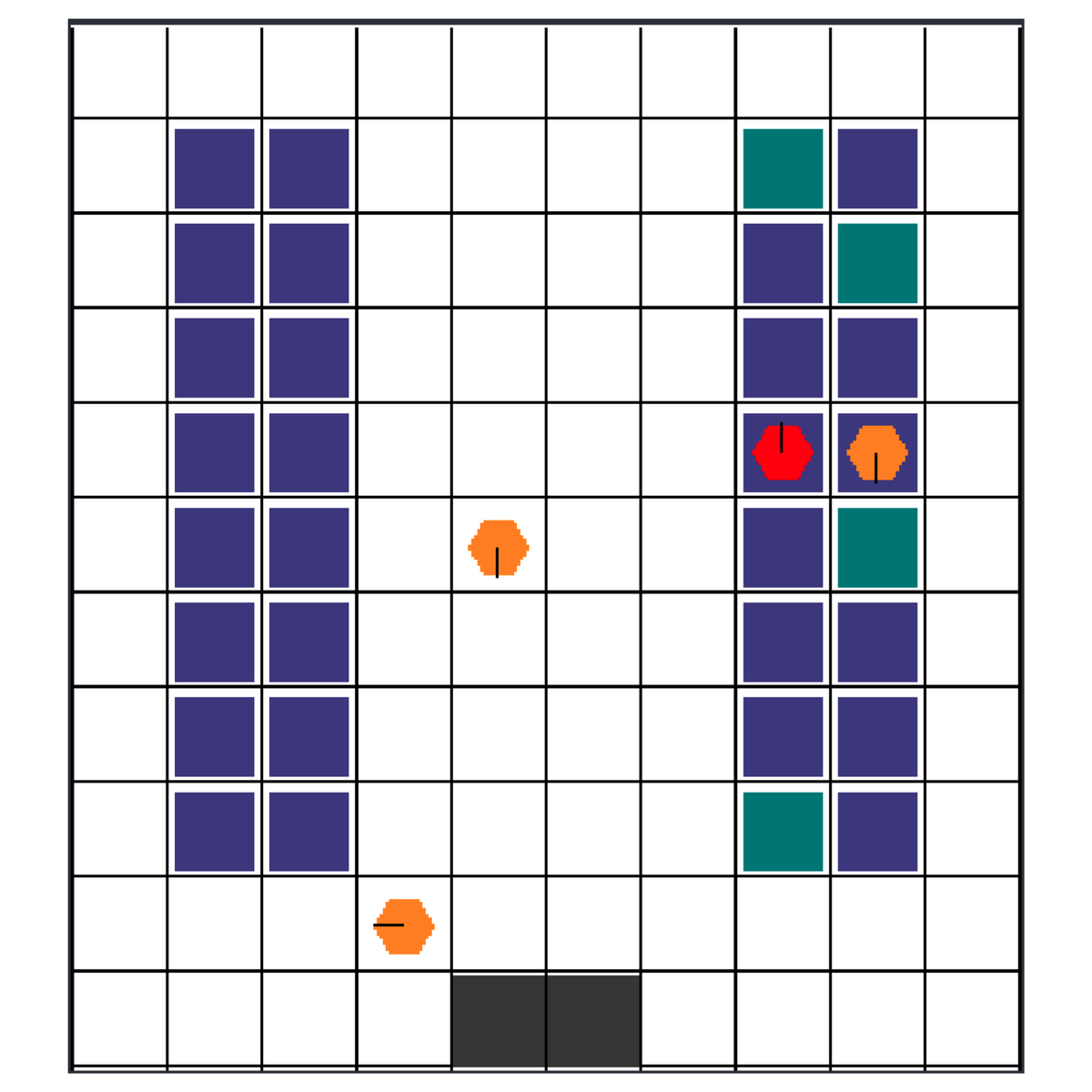}\hfill
    \includegraphics[width=0.15\linewidth]{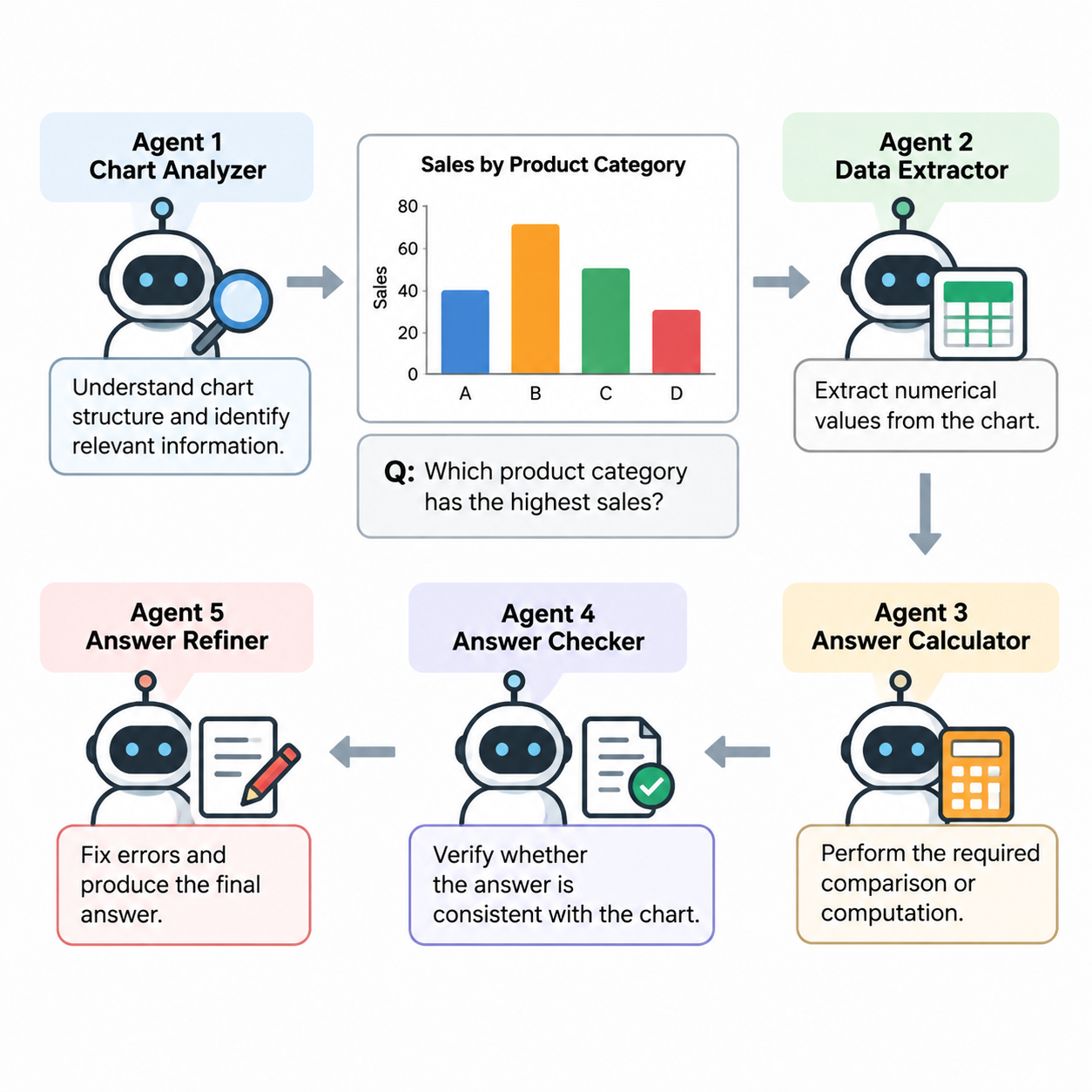}\hfill
    \includegraphics[width=0.15\linewidth]{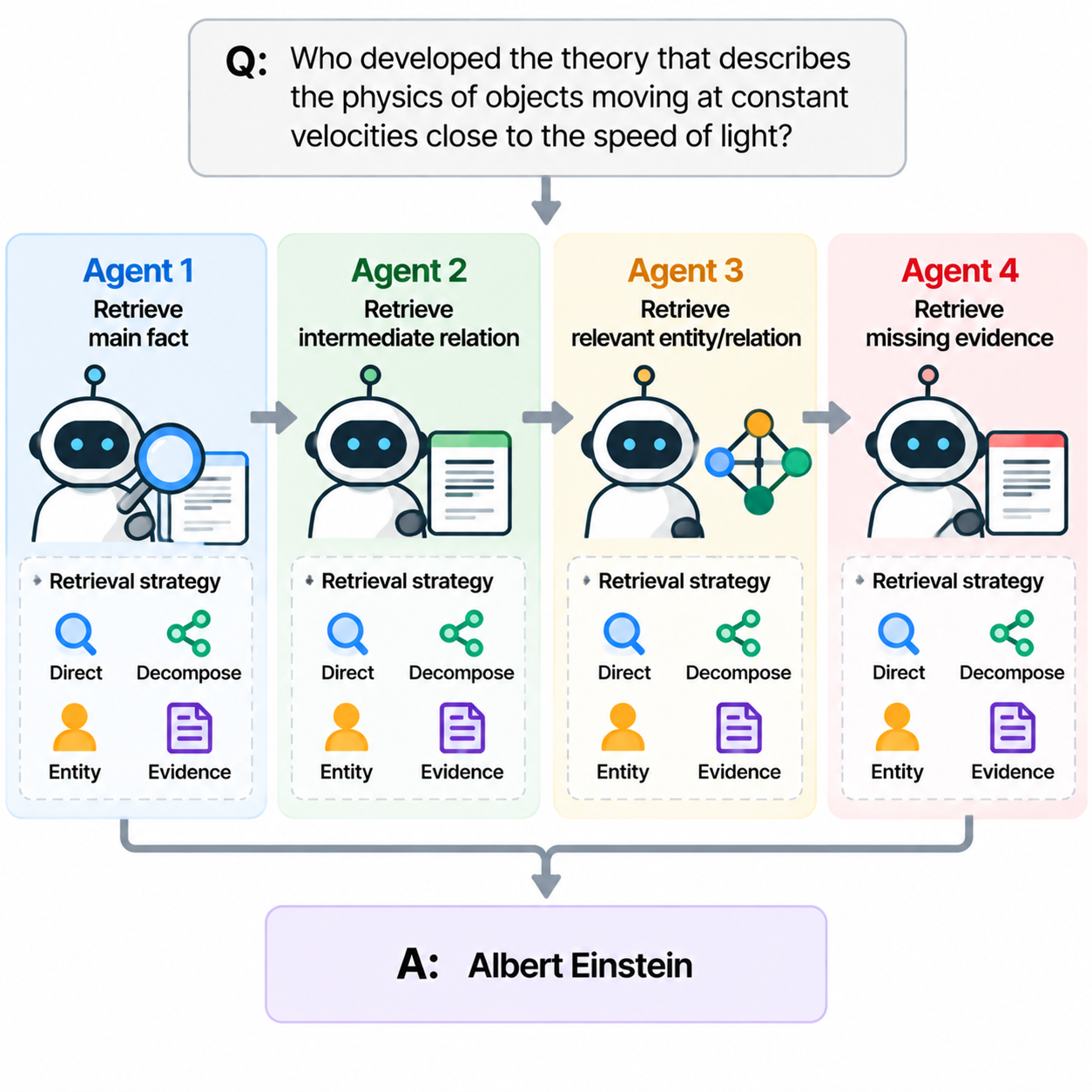}

    \par\vspace{3pt}
    \noindent
    \makebox[0.15\linewidth][c]{\small (a) Spread}\hfill
    \makebox[0.15\linewidth][c]{\small (b) Speaker--Listener}\hfill
    \makebox[0.15\linewidth][c]{\small (c) LBF}\hfill
    \makebox[0.15\linewidth][c]{\small (d) RWARE}\hfill
    \makebox[0.15\linewidth][c]{\small (e) ChartQA}\hfill
    \makebox[0.15\linewidth][c]{\small (f) MuSiQue}
    % \vspace{-2mm}
    \caption{Illustrations of the selected MARL benchmarks for testing LOD-MAPPO.}
    \label{fig:testbeds}
\end{figure}

\begin{figure}[b!]
    \centering
    \begin{subfigure}
        \centering
        \includegraphics[width=0.22\textwidth]{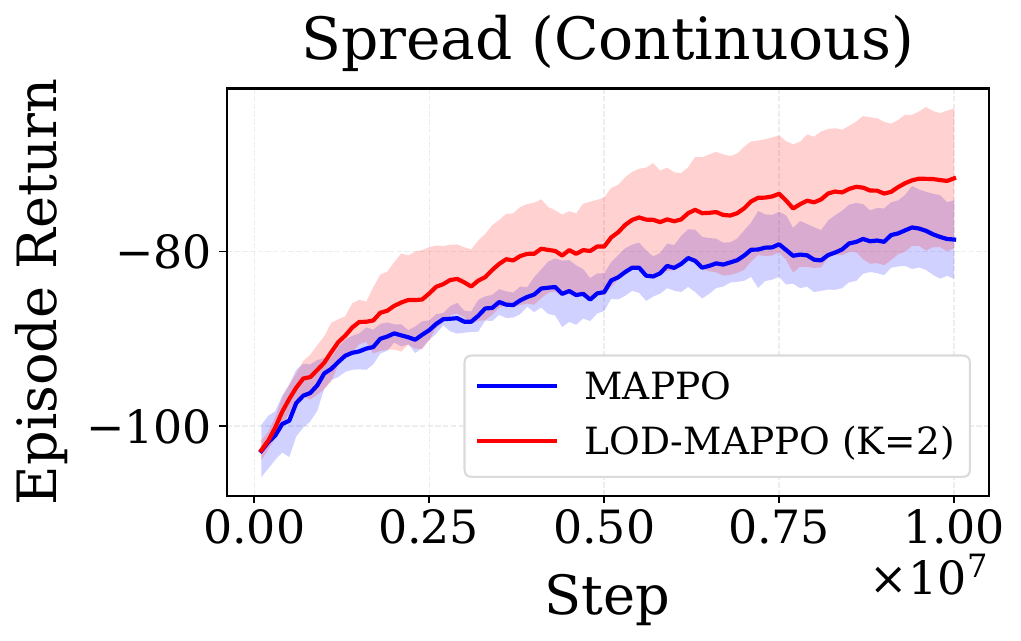}
        % \caption{Continuous setting}
        % \label{fig:continuous}
    \end{subfigure}
    \hfill
    \begin{subfigure}
        \centering
        \includegraphics[width=0.22\textwidth]{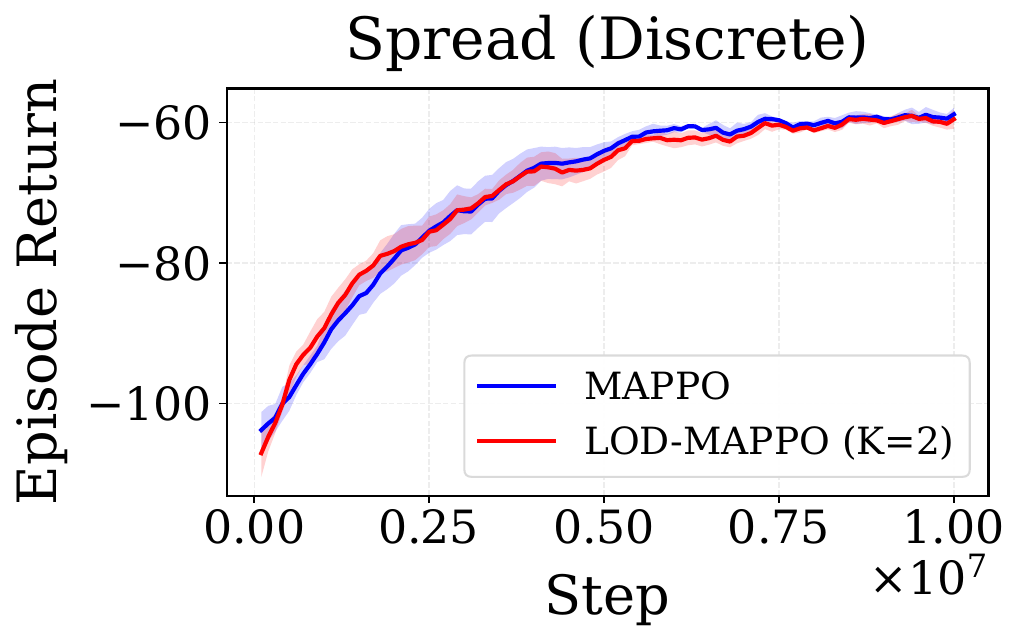}
    \end{subfigure}
    \begin{subfigure}
        \centering
        \includegraphics[width=0.22\textwidth]{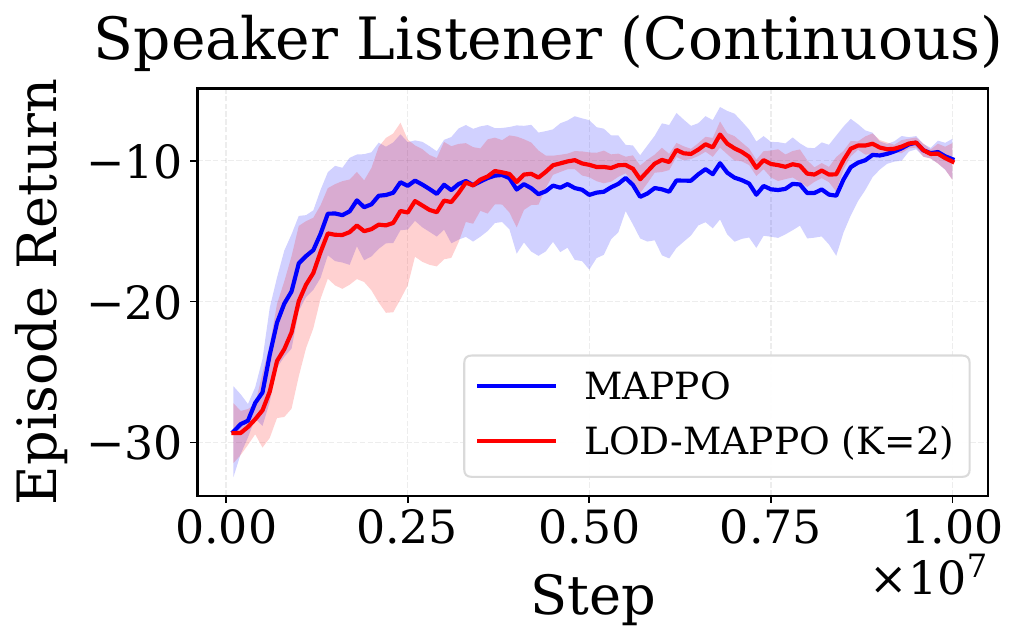}
    \end{subfigure}
    \hfill
    \begin{subfigure}
        \centering
        \includegraphics[width=0.22\textwidth]{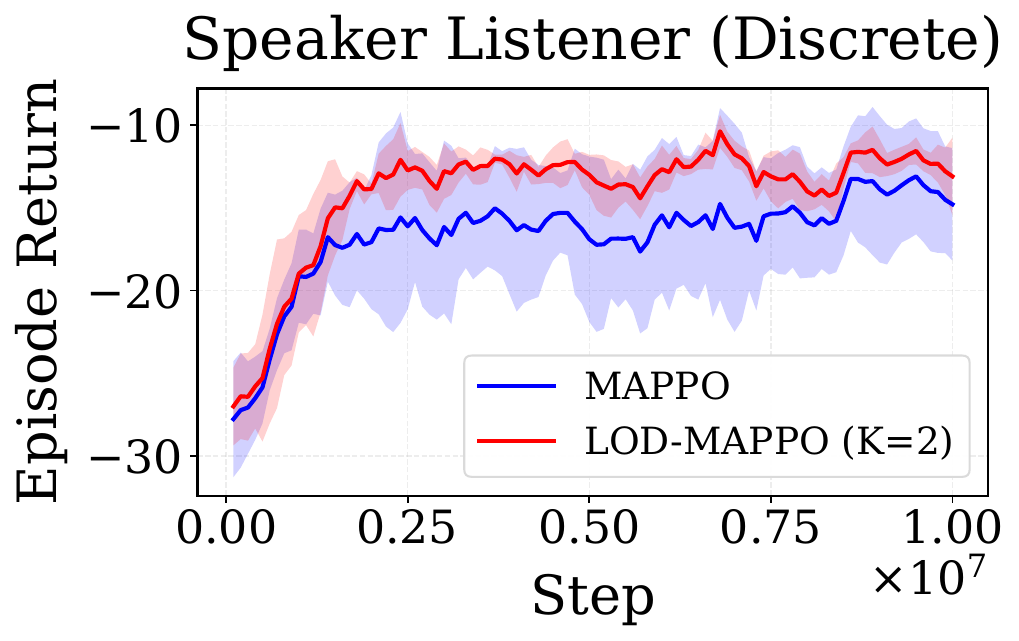}
    \end{subfigure}
    \vspace{-5mm}
    \caption{Comparison between LOD-MAPPO~($K=2$) and MAPPO in two Multi-Agent Particle Environments~(MPE) across continuous and discrete action settings over five random seeds.}
    \label{fig:MPE}
\end{figure}

\subsection{Multi-Agent Particle Environment~(MPE)}
% github in \citep{christianos2021scaling}

\noindent \textbf{Environmental Setup.} We consider two fully cooperative tasks in MPE across continuous and discrete action settings: \texttt{simple\_spread} and \texttt{Speaker\_listener}. (1) \texttt{simple\_spread:} a cooperative game with $N$ agents to cover $N$ landmarks without collision, which implies two intrinsic objectives. We choose $N=3$ and $K=2$ in the main experiment and we also perform sensitivity analysis across various $N$ and $K$; (2) \texttt{Speaker\_listener:} one static speaker has to communicate a goal landmark to a listener capable of moving. We choose $N=2$ and $K=2$. Our implementations are adapted from the widely adopted HARL codebase from \citep{zhong2024heterogeneous}.

\noindent \textbf{Main Results.} Figure~\ref{fig:MPE} showcases that LOD-MAPPO outperforms MAPPO across two MPE tasks with both continuous and discrete action spaces, except for Spread~(Discrete), where LOD-MAPPO is on par with MAPPO. In  \texttt{simple\_spread}, the reward combines a shared coverage component with agent-specific collision penalties. Our LOD method can exploit this structure by learning latent reward factors and reconstructing each agent’s advantage through agent-specific mixing weights, which tend to  improve value estimation and policy optimization. In  \texttt{Speaker\_listerner}, the speaker and the listener have complementary roles. The intermediate-stage gains in the continuous setting and the more persistent improvement in the discrete setting therefore demonstrate an empirical benefit of LOD. In Figure~\ref{fig:MPE_ablation} of Appendix~\ref{appendix:experiment_MPE}, we vary $N$ between 4 and 6, and a wide range of $\tau$ used in the soft updating of $\Gamma$ on Spread~(Continuous). The benefits of LOD-MAPPO are similar and further discussion is provided in Appendix~\ref{appendix:experiment_MPE}.

% \clearpage
\subsection{Level-Based Foraging~(LBF) and Multi-Robot Warehouse~(RWARE)}

\noindent \textbf{Environmental Setup.} We evaluate LOD-MAPPO on Level-Based Foraging (LBF) and the Multi-Robot Warehouse (RWARE), two environments requiring coordination under sparse rewards. (1) \texttt{Level-Based Foraging (LBF):} agents navigate a grid and collect items when their combined levels meet or exceed the item’s level. Rewards are heterogeneous across agents, and the cooperative variant (denoted by \texttt{LBF-Coop}) requires joint collection, further increasing the difficulty of obtaining reward. We consider both variants on a $15\times15$ grid. (2) \texttt{Multi-Robot Warehouse (RWARE):} partially observable robots must collect requested shelves and deliver them to designated locations. Rewards occur only upon successful delivery, requiring extended sequences of coordinated actions. We evaluate the tiny and small layouts. All tasks use $N=4$ agents, and results are averaged over five random seeds. Our implementations are directly adapted from the EpyMARL framework~\citep{papoudakis2021Benchmarking} with the same hyperparameters as MAPPO.
% \vspace{-5mm}
\begin{figure}[htbp]
    \centering
    \begin{subfigure}
        \centering
        \includegraphics[width=0.22\textwidth]{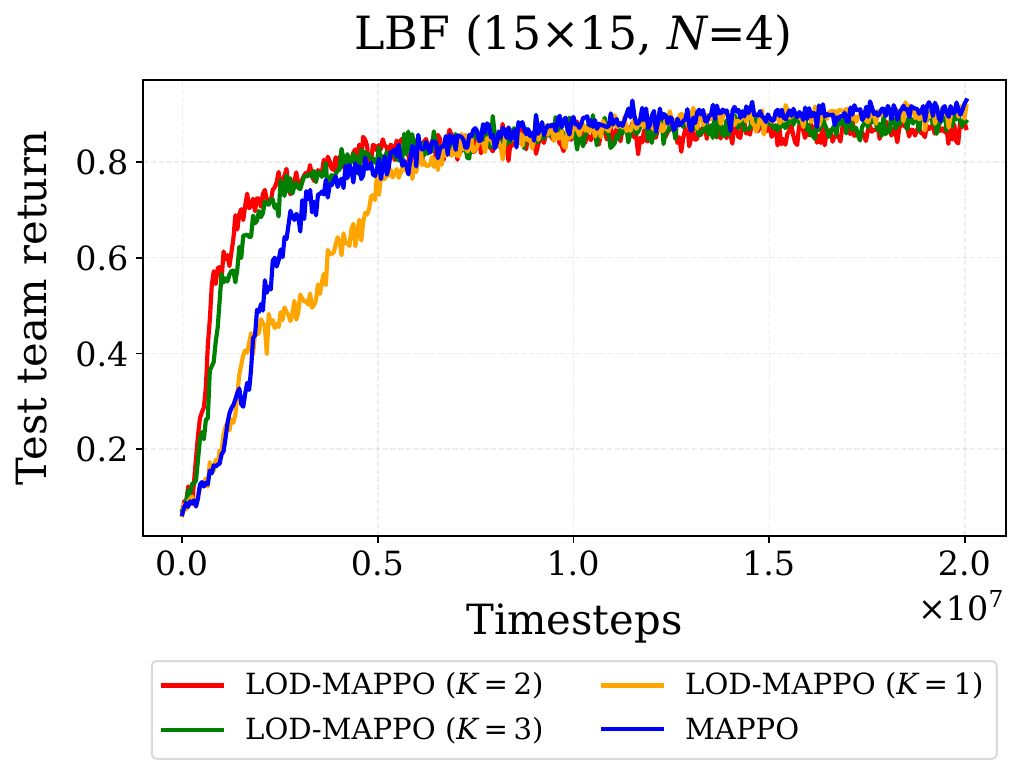}
    \end{subfigure}
    \hfill
    \begin{subfigure}
        \centering
        \includegraphics[width=0.22\textwidth]{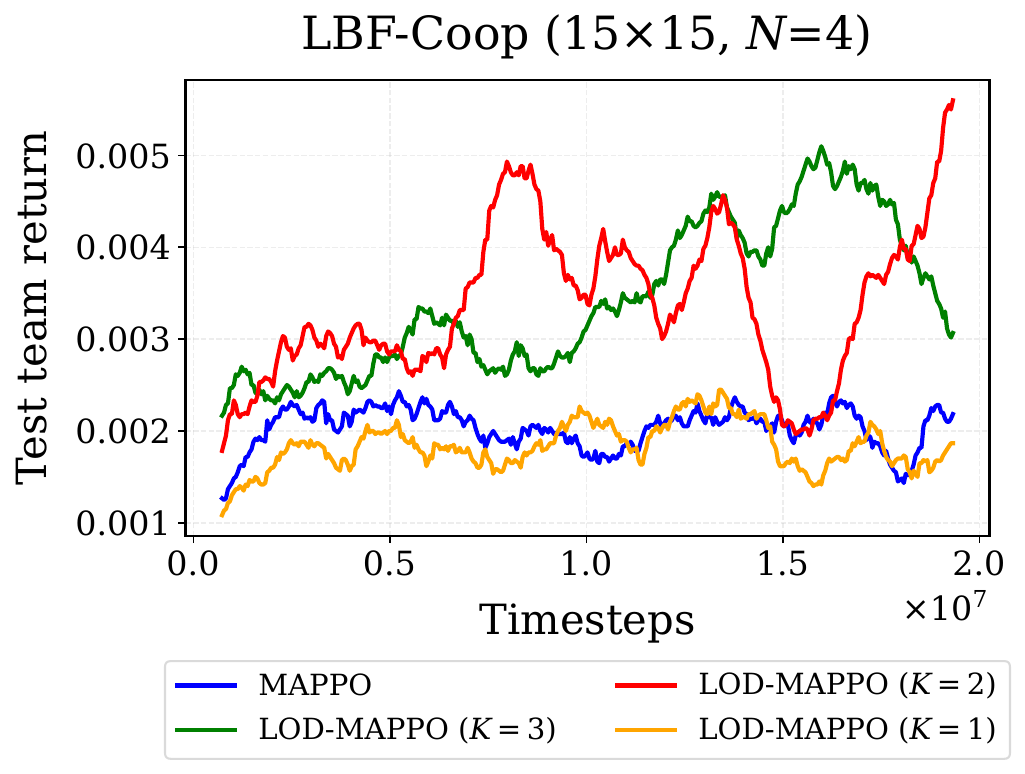}
    \end{subfigure}
    \begin{subfigure}
        \centering
        \includegraphics[width=0.22\textwidth]{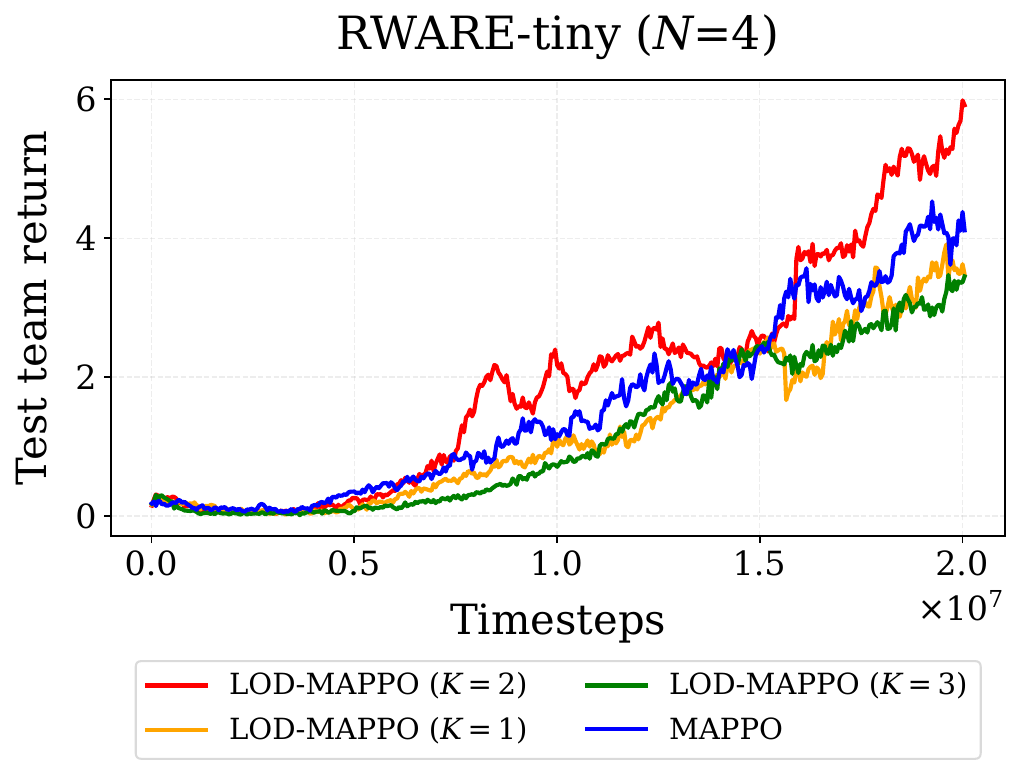}
    \end{subfigure}
    \hfill
    \begin{subfigure}
        \centering
        \includegraphics[width=0.22\textwidth]{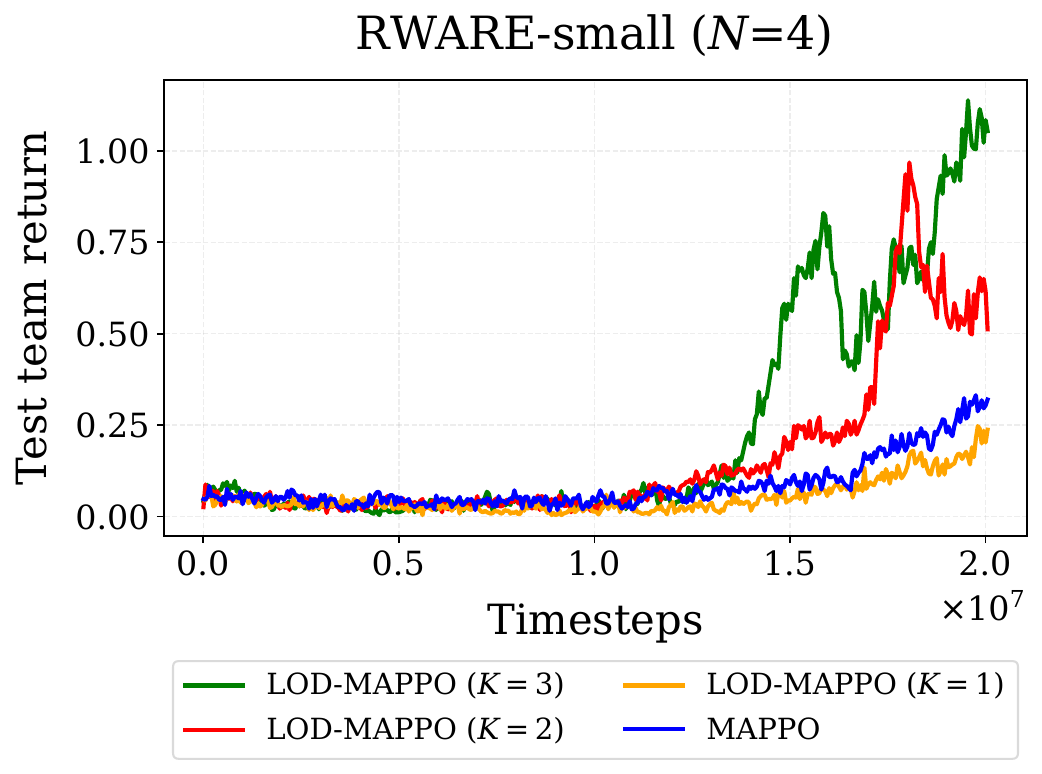}
    \end{subfigure}
    \vspace{-5mm}
    \caption{Comparison between LOD-MAPPO across a wide range of $K$ and MAPPO in LBF and RWARE environments. The learning curves are the mean returns across five random seeds.}
    \label{fig:LBF_RWARE}
\end{figure}

\noindent \textbf{Main Results.} Figure~\ref{fig:LBF_RWARE} shows the advantages of LOD-MAPPO over MAPPO, where the learning curves are the mean averaged across five random seeds to facilitate comparison. In standard LBF, LOD-MAPPO with $K=2$ or $K=3$ improves more rapidly during early training, while all methods eventually attain similar returns. In cooperative LBF, LOD-MAPPO with $K=2$ or $K=3$ achieves higher returns over substantial portions of training than MAPPO. Nevertheless, absolute returns remain low and the curves fluctuate considerably, indicating that sparse cooperative rewards remain a major challenge. In RWARE-tiny and RWARE-small, LOD-MAPPO with $K=2$ and $K=3$ achieves the highest final mean return, respectively. 
% In RWARE-tiny, LOD-MAPPO with $K=3$ and $K=1$ finish below MAPPO, while in RWARE-small, LOD-MAPPO with $K=2$ and $K=3$ significantly outperforms MAPPO and $K=1$. 
These results suggest that multiple latent reward factors can provide useful structure for learning from sparse, agent-dependent reward signals. By sharing latent value or advantage estimates while retaining agent-specific mixing weights, LOD may facilitate the use of limited reward information across agents. In addition, the optimal choice of $K$ is task-dependent and may require task-specific tuning. The learning curves with shaded bands indicating variations are provided in Figure~\ref{fig:LBF_RWARE_shade} in Appendix~\ref{appendix:experiment_LBFRWARE}.

% \clearpage
\subsection{Agentic RL Settings}\label{sec:experiment_agenticRL}

\noindent \textbf{Environmental Setup.}  In agentic RL, the environment that agents interact with often encodes a shared and sequence-level reward, even though they may perform distinct roles and contribute differently to the final outcome. Therefore, we deploy the local advantage alternative via Eq.~\ref{eq:local_advantage} to characterize objective heterogeneity for our latent objective discovery approach. We consider \texttt{ChartQA} and \texttt{MuSiQue Multi-Hop Retrieval} as two agentic
RL tasks from Agent Lightning~\citep{masry2022chartqa,he2026agent} and use Qwen2.5~\citep{hui2024qwen2} in policy learning.  Agents occupy distinct functional roles and make
heterogeneous strategy decisions, naturally motivating LOD to uncover the latent objectives shared across all agents.  All agents receive a shared binary task reward, termed
\emph{Success}. In \texttt{ChartQA}, the reward is 1 if the workflow successfully produces the final answer after checking and refinement
and zero otherwise, while in \texttt{MuSiQue}, success indicates the required supporting information
is successfully retrieved and zero otherwise. We report the moving-average success rate during training. More experimental details are provided in Appendix~\ref{appendix:experiment_agenticRL}.
\begin{enumerate}[label=\textcircled{\arabic*},itemindent=\parindent,leftmargin=*, topsep=0pt, partopsep=0pt,itemsep=2pt,parsep=2pt]
   \item \textbf{ChartQA:} The workflow consists of five successive roles: \emph{Chart Analyzer}, which interprets the chart; \emph{Data Extractor}, which extracts the relevant values; \emph{Answer Calculator}, which performs the required computation; \emph{Answer Checker}, which verifies the result; and \emph{Answer Refiner}, which produces the final answer after correction. We treat these five roles as cooperating heterogeneous agents and each role selects among four role-specific strategies to make different types of decisions while jointly determining the task outcome.

   \item \textbf{MuSiQue Multi-Hop Retrieval.} It consists of questions that require combining information from multiple sources. The workflow consists of four successive retrieval steps, which we treat as cooperating agents. At each step, the corresponding agent selects among four retrieval strategies: \emph{Direct}, which searches for the main fact; \emph{Decompose}, which targets an unresolved intermediate relation; \emph{Entity}, which focuses on a relevant entity and relation; and \emph{Evidence}, which searches for information missing from the
available evidence. The selected strategy is used to generate a search
query and retrieve supporting information. 
\end{enumerate}

% Across both tasks, yet are optimized under the same
% task-level reward. This mismatch between heterogeneous agent objectives
% and homogeneous reward supervision naturally motivates LOD to uncover
% the latent objectives underlying their different contributions.

\begin{figure}[htpb]
    \centering
    \includegraphics[width=0.4\linewidth, trim={10pt 40pt 10pt 10pt}, clip]{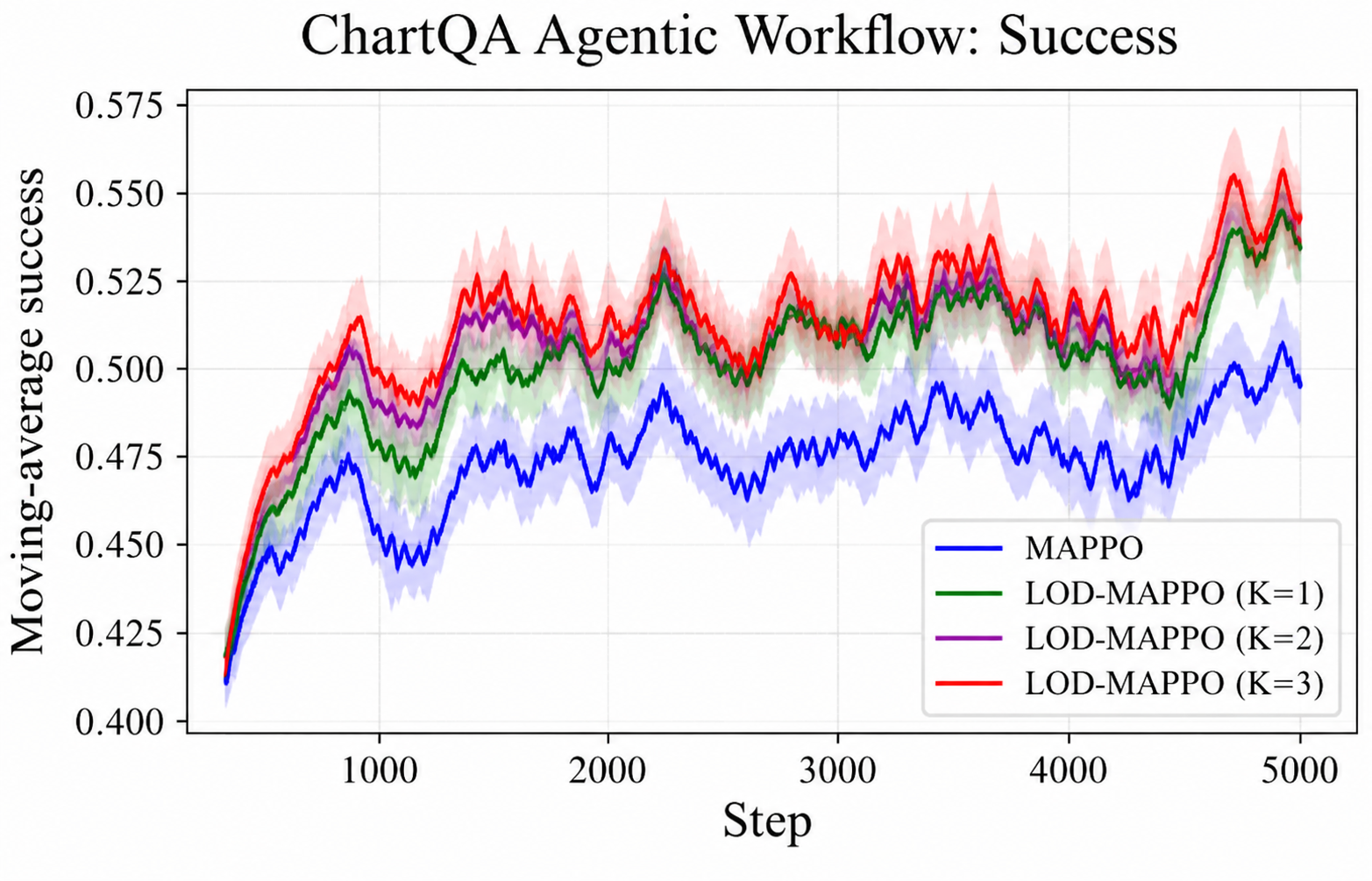}
    \hspace{0.03\linewidth}
    \includegraphics[width=0.4\linewidth, trim={10pt 40pt 10pt 10pt}, clip]{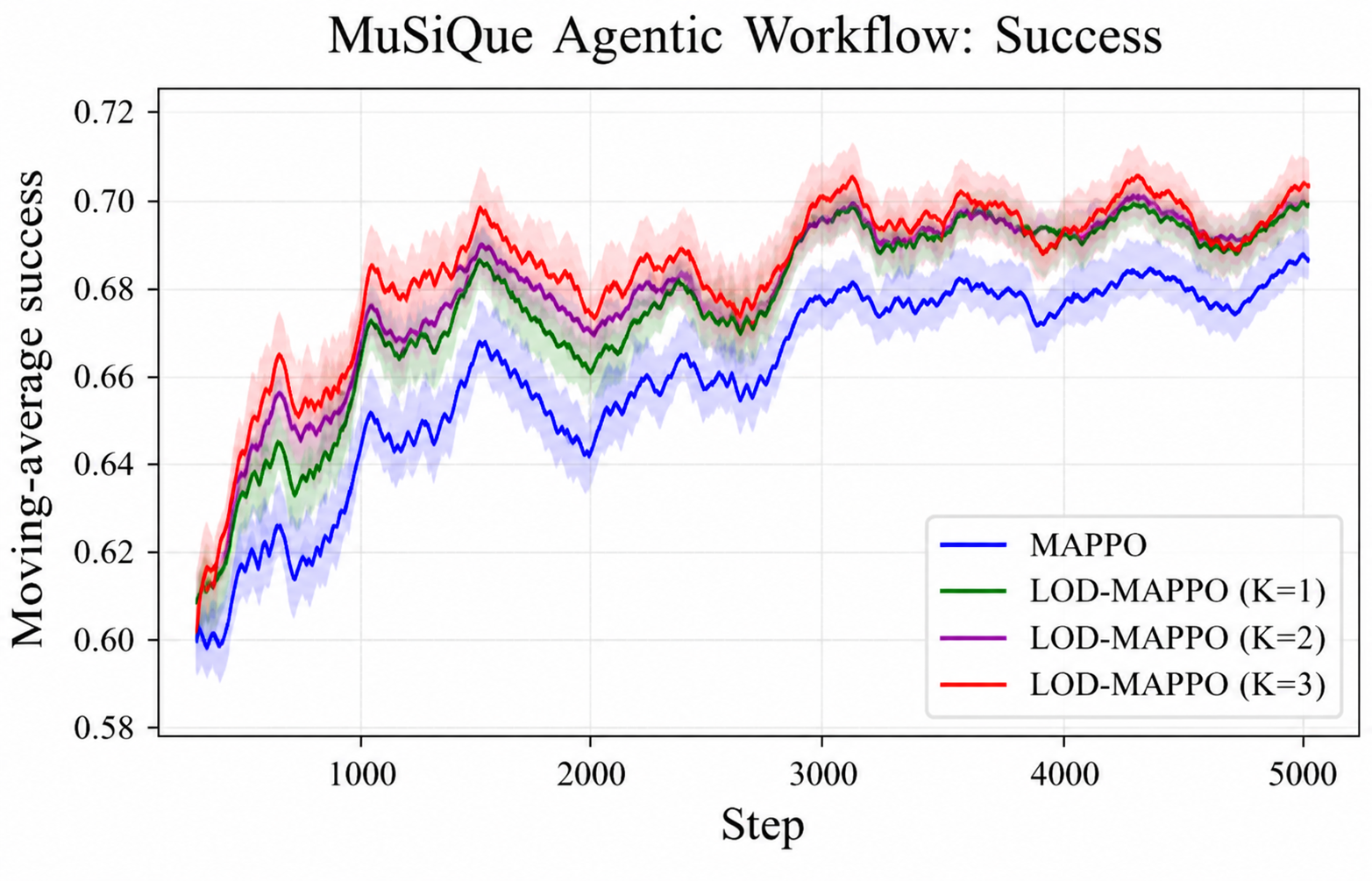}
    \vspace{-2mm}
    \caption{Comparison between LOD-MAPPO and MAPPO in two Agent Lightning environments.}
    \label{fig:agent_lightning}
\end{figure}

\noindent \textbf{Main Results.} Figure~\ref{fig:agent_lightning} shows that LOD-MAPPO with diverse $K$ exhibits faster improvement during the early
training stage, indicating that learning the latent structure via LOD benefits the training when agents are still learning their roles
under a shared reward. As training progresses, LOD-MAPPO maintains a consistent advantage over
MAPPO, suggesting that latent objective learning continues to
support coordination among heterogeneous agents. Increasing the number of
latent objectives generally leads to better performance, but the benefit gradually saturates as the policies converge. The optimal choice of $K$ may depend on the task, which requires prior knowledge.

\section{Conclusion and Limitations}

In this study, we introduce latent objective discovery as a new MARL framework that extends existing CTDE and networked MARL formulations by adaptively uncovering shared latent structure in agents’ heterogeneous objectives. Inspired by the mixed membership concept, the framework learns shared latent objectives through multiple mixed centralized critics and captures each agent’s relationship to these objectives through personalized weights.  These weights offer an interpretable characterization of agents’ objective preferences and support personalized policy learning in broader MARL settings.

Although the framework is, in principle, applicable to cooperative, competitive, and mixed settings~\citep{amato2024introduction}, our empirical evaluation focuses on cooperative tasks. Evaluating its effectiveness in competitive and mixed settings and establishing theoretical guarantees under nonlinear function approximation remain important directions for future work. Our study also focuses specifically on objective heterogeneity and extending latent objective discovery to accommodate other forms of heterogeneity in MARL warrants further investigation.
% The reward factor model often assume the stationarity, but this assumption may not necessarily satisfied in the online RL algorithms, where the variation of the reward time series depends on the RL algorithm that is constantly updated in our deployment. The rigorous convergence is an interesting theoretical research problem in the future.

\section*{Ethics Statement}

This work studies multi-agent reinforcement learning in simulated environments. Improved coordination does not guarantee safe or equitable outcomes. Real-world deployment requires careful reward design and evaluation of safety, robustness, and impacts on the interaction of agents' behaviors in safety-critical scenarios.

\section*{Reproducibility Statement}

Our implementations are directly adapted from the widely adopted MARL frameworks, including HARL~\citep{zhong2024heterogeneous} and EpyMARL~\citep{papoudakis2021Benchmarking}. Detailed experimental descriptions are provided in Appendix~\ref{appendix:experiment} to facilitate reproducibility. The code is also submitted in the supplementary materials.

\section*{Statement on the Use of AI}
In this work, we used generative AI tools to help guide and refine the experiment implementations and polish the writing of this study.  We take responsibility for the final content of this work, including text, claims or artifacts produced with the aid of generative AI.

\bibliography{reference}
\bibliographystyle{iclr2027_conference}

\clearpage
\appendix
%%%%%%%%%%%% set table of content for appendix (start)
\addcontentsline{toc}{section}{Appendix} % Add the appendix text to the document TOC
\part{Appendix} % Start the appendix part
\parttoc % Insert the appendix TOC
%%%%%%%%%%% set table of content for appendix (end)
\clearpage
\section{Related Work}\label{app:relatedwork}

\textbf{MARL with Heterogeneous Agents.} Most MARL algorithms, including value factorization~\citep{sunehag2017value,rashid2020monotonic,wang2020qplex} and centralized critics~\citep{lowe2017multi,foerster2018counterfactual,yu2022surprising}, assume a shared global reward function in a cooperative setting and inherently imply homogeneous agent behaviors. To enable diverse and specialized behaviors, parameter sharing with diversity encouraging~\citep{li2021celebrating} and specialized fine-tuning~\citep{terry2020revisiting, christianos2021scaling, li2024adaptive,kim2023parameter,zhang2025low,hu2025heterogeneity} are proposed, but they often possess limited applicability~\citep{hu2022policy,zhong2024heterogeneous}. \citet{christianos2021scaling} recognizes the behavior heterogeneity through hard clustering in a latent trajectory space to apply the selective parameter sharing. By 
contrast, our approach focuses on discovering latent and possibly overlapping objectives specifically in the reward space, capturing heterogeneity at the level of incentives beyond behavior similarity. \citet{kuba2021trust, zhong2024heterogeneous} performed a sequential update scheme based on the advantage decomposition to account for the heterogeneous agents in the regime with a shared reward function, but they did not explicitly define or characterize heterogeneity of policies among agents and only focused on the cooperative setting. 

\textbf{Latent Representation in MARL.} \citet{na2024lagma,na2025trajectory} propose to generate a goal-reaching trajectory in latent state space to provide a latent goal-guided incentive or identify the class of trajectories in latent trajectory space to perform task-aware execution. Task division methods, such as role-based task factorization methods~\citep{wang2020roma,wang2020rode,tennenholtz2019natural,chandak2019learning}, subtask assignment~\citep{yang2022ldsa}, and skill learning~\citep{liu2022heterogeneous}, factor the agent's identity or action space to define different subtasks, skills, or roles in latent spaces. In contrast, our latent objective discovery operates the factorization at complementary levels of abstraction: the former structures how agents act, while the latter structures why agents act.

\textbf{Networked MARL.} Networked MARL~\citep{varshavskaya2009efficient,zhang2018fully,zhang2021multi,lin2021multi} is typically seen as a theoretical framework that connects heterogeneous agents with individual reward functions through a graph and executes decentralized control by leveraging the decentralized information. Although this paradigm naturally contributes to local privacy and communication efficiency in network control~\citep{qu2020scalable,chu2020multi,yi2022learning,du2022scalable,nayak2023scalable,ma2024efficient}, the network topology above is often explicitly pre-specified, heuristically designed, or externally observed, restricting coordination to hard-coded local structure. By contrast, our latent objective discovery induces a latent, non-local, and statistically principled network topology to learn objective abstractions guided by agents' diverse incentives. 
% Exponential decay when applied in Q-function~\citep{lin2021multi} is also able to induce non-local network topology in networked MARL, whereas our hidden network is constructed to capture the hidden reward structure.

% By utilizing the low-rank approximation, factorized Q function~\citep{zhou2019factorized} was proposed to allow parameter sharing among agents within the same group in multi-agent systems. 

% Grouping was also probed in large-scale traffic signal control~\citep{liu2023gplight} based on the spatialtemporal features of agents' environments. 

% \noindent \textbf{Multi-agent Interactions in Agentic RL.} \ke{to be added}

\clearpage

\section{Theoretical Results}\label{appendix:theory}
\subsection{Proof of Proposition~\ref{prop:critic_decomposition}}\label{app:proof_prop_criticdecompose}

\begin{proof}
    For the agent $i$, the latent reward decomposition in Eq.~\ref{eq:reward_decomposition} indicates that
    \begin{align*}
        R_t^i = \gamma_i^\top \boldsymbol{F}_t + \epsilon_t^i.
    \end{align*}
    Recall the definition of the action-value function $Q^{\boldsymbol{\pi}}_i(s, \boldsymbol{a})$ for each agent $i$:
    \begin{align*}
        Q^{\boldsymbol{\pi}}_i(s, \boldsymbol{a}) = \mathbb{E}_\pi\left[\sum_{t=0}^{\infty} \gamma^{t} R_{t}^i \mid s_0=s, a_0=\boldsymbol{a}\right].
    \end{align*}
    Next, we plug the equation of $R_t^i$ into the definition of $Q^{\boldsymbol{\pi}}_i$ above, which results in
    \begin{align*}
         Q^{\boldsymbol{\pi}}_i(s, \boldsymbol{a}) & \overset{(a)}{=} \gamma_i^\top  \mathbb{E}_\pi\left[\sum_{t=0}^{\infty} \gamma^{t} \boldsymbol{F}_{t} \mid s_0=s, a_0=\boldsymbol{a}\right] + \mathbb{E}_\pi\left[\sum_{t=0}^{\infty} \gamma^{t} \epsilon_{t}^i \mid s_0=s, a_0=\boldsymbol{a}\right] \\
         & \overset{(b)}{=} \gamma_i^\top  \mathbb{E}_\pi\left[\sum_{t=0}^{\infty} \gamma^{t} \boldsymbol{F}_{t} \mid s_0=s, a_0=\boldsymbol{a}\right] \\
         & := \gamma_i^\top \boldsymbol{Q}_{\boldsymbol{F}}^{\boldsymbol{\pi}}(s, \boldsymbol{a}),
    \end{align*}
    where $(a)$ is true due to the fact that $\Gamma$ is time-independent and $(b)$ holds because $\epsilon_t^i$ is the white noise  and $\mathbb{E}_\pi\left[ \epsilon_{t}^i \mid s_0=s, a_0=\boldsymbol{a}\right] =0$. By the definition of $\boldsymbol{Q}_{\boldsymbol{F}}$, the simplification above holds, which further implies that
     \begin{align}
            \boldsymbol{Q}^\pi(s, \boldsymbol{a}) = \Gamma  \boldsymbol{Q}_{\boldsymbol{F}}^{\boldsymbol{\pi}}(s, \boldsymbol{a}).
        \end{align}
    
\end{proof}

\subsection{Proof of Theoretical Convergence}\label{app:theorydetails}

In Theorems \ref{thm:convergen_omega} and \ref{thm:convergence_theta}, we provided the theoretical guarantee for our proposed generic actor-critic algorithm (See Algorithm~\ref{alg:generic_averagereward} in Appendix~\ref{appendix:algorithm_averagerewards}). In this section, we present more details on the theoretical analysis, including the full mathematical definitions of the equations appeared in these theorems.

To this end, we introduce some additional notations for our theoretical results. With a slight abuse of the notation, let $P^{\theta}(s',a'|s,a)=P(s'|s,a)\pi_{\theta}(s',a')$. Here, it should be distinguished within context easily from the transition probability matrix $P^{\theta}(s'|s)$ of the Markov chain $\{s_t\}_{t\ge 0}$ under the policy $\pi_{\theta}$. Let $D_{\theta}^{s,a}=\text{diag}(d_{\theta}(s)\pi_{\theta}(s,a),s\in\mcal{S},a\in\mcal{A})$. For each factor $k$, let $R^{(k)}=[R^{(k)}(s,a),s\in\mcal{S},a\in\mcal{A}]'\in\mbb{R}^{|\mcal{S}||\mcal{A}|}$ and $J^{(k)}(\theta)=\sum_{s\in\mcal{S},a\in\mcal{A}}d_{\theta}(s)\pi_{\theta}(s,a)R^{(k)}(s,a)$, for $1\le k\le K$. They represent the $k$-th latent factor of the objectives reward $R^i$ and $J(\theta)$ induced by the factor model \eqref{eq:reward_decomposition}. We also define an operator $T_{\theta}^{Q,(k)}:\mbb{R}^{|\mcal{S}||\mcal{A}|}\rightarrow \mbb{R}^{|\mcal{S}||\mcal{A}|}$ for any action-value vector $Q\in \mbb{R}^{|\mcal{S}||\mcal{A}|}$ as
\begin{align*}
    T_{\theta}^{Q,(k)}(Q)=R^{(k)}-J^{(k)}(\theta)\cdot\mbf{1}+P^{\theta}Q,\ 1\le k\le K.
\end{align*}

Now, we first provide the details of Theorem \ref{thm:convergen_omega}.
\begin{Theorem}\label{appthm:convergen_omega}
    Under Assumption \ref{Assump_1} and \ref{Assump_3} to \ref{Assump_5}, for any policy $\pi_{\theta}$ given, the factor parameter $\omega_{t}^{(k)}$ converges: $\lim_t\omega_{t}^{(k)}=\omega_{\theta}$ almost surely and the limit $\omega_{\theta}$ is the unique solution to
    \begin{align*}
        \Phi'D_{\theta}^{s,a}(T_{\theta}^{Q,(k)}(\Phi\omega_{\theta}^{(k)})-\Phi\omega_{\theta}^{(k)})=0,\ 1\le k\le K.
    \end{align*}
\end{Theorem}

Next we provide the details of Theorem \ref{thm:convergence_theta}. During the actor step, in order to show the convergence of the update in $\theta$, we introduce:
\begin{align}\label{Apsi}
    A_{t,\theta}^i&=\phi_t'\omega_{\theta}^i-\sum_{a^i\in \mcal{A}}\pi_{\theta^i}^i(s_t,a^i)\phi(s_t,a^i,a_t^{-i})'\omega_{\theta}^i,\nonumber\\
    \psi_{t,\theta}^i&=\nabla_{\theta^i}\log\pi_{\theta^i}^i(s_t,a_t^i),
\end{align}
where $\phi_t=\phi(s_t,a_t)$ and $\omega_{\theta}^i$ is defined as follows. Since the actor step is updating $\theta$ for each agent's policy $\pi_{\theta^i}^i$ and we already defined the  limiting factor parameter $\omega_{\theta}^{(k)}$ in Theorem \ref{appthm:convergen_omega}, the limiting agent parameter $\omega_{\theta}^i$ is thus defined by enforcing the factor model structure $\omega_{\theta}^i=\sum_{k=1}^{K}\gamma_i(k)\omega_{\theta}^{(k)}$. Note here that due to the linear assumption of the $Q$-function, it suffices to only linearly aggregate its parameter $\omega$ (otherwise, we need to strictly aggregate $Q$-functions instead). Furthermore, for each $i\in\mcal{N}$, define a vector 
\begin{align*}
    \mcal{L}^i(g(\theta)):=\lim_{\eta\rightarrow 0^+}\frac{\mcal{L}_0^i(\theta^i+\eta g(\theta))-\theta^i}{\eta},
\end{align*}
for any $\theta\in\Theta$ and $g:\Theta\rightarrow \mbb{R}^{\sum_{i\in\mcal{N}}m_i}$ a continuous function. If the above limit is not unique, we define $\mcal{L}^i(g(\theta))$ to be the set of all possible limit points of the right-hand side. The details of Theorem \ref{thm:convergence_theta} are included in the following theorem.
\begin{Theorem}
    Under Assumption \ref{Assump_1} to Assumption \ref{Assump_5}, the agent policy parameter $\theta_t^i$ converges almost surely to a point in the set of the asymptotically stable equilibria of
    \begin{align*}
        \dot{\theta}^i=\mcal{L}^i(\mbb{E}_{s_t\sim d_{\theta},a_t\sim \pi_{\theta}}(A_{t,\theta}^i\psi_{t,\theta}^i)),\ i\in\mcal{N}.
    \end{align*}
\end{Theorem}

\subsection{Auxiliary Results}
\subsubsection{Results on Stochastic Approximation}
In this section, we introduce a basic result of stochastic approximation shown in \cite{borkar2008stochastic}. More general conclusions can also be found in \cite{benaim2006dynamics,harold1997stochastic}.

Consider the following $n$-dimensional ($n\in\mbb{N}$) stochastic approximation iteration:
\begin{align*}
    x_{t+1}=x_t+\zeta_t(h(x_t,Y_t)+M_{t+1}+\eta_{t+1}),\ t\ge 0,
\end{align*}
where $\zeta_t>0$ and $\{Y_t\}_{t\ge 0}$ is a Markov chain on a finite set $A$. 

\begin{Theorem}\label{supp1}
    Suppose the following assumptions hold:
    \begin{itemize}
        \item [(1).] $h:\mbb{R}^n\times A\rightarrow \mbb{R}^n$ is Lipschitz continuous in its first argument;
        \item [(2).] $\{Y_t\}_{t\ge 0}$ is an irreducible Markov chain with stationary distribution $\pi$;
        \item [(3).] The stepsize sequence $\zeta_t$ is such that $\sum_t\zeta_t=\infty$ and $\sum_t\zeta_t^2<\infty$;
        \item [(4).] $\{M_t\}$ is a martingale difference sequence, i.e., $\mbb{E}(M_{t+1}|x_\tau,M_\tau,Y_\tau,\tau\le t)=0$, and it holds that for some $C>0$ and $t\ge 0$,
        \begin{align*}
            \mbb{E}\left(\|M_{t+1}\|^2|x_\tau,M_\tau,Y_\tau,\tau\le t\right)\le C(1+\|x_t\|^2);
        \end{align*}
        \item [(5).] The sequence $\{\eta_t\}$ is a bounded random sequence with $\lim_t\eta_t=0$ almost surely.
    \end{itemize}
    Then the asymptotic behavior of the above iteration is associated with the behavior of the solution to the ODE:
    \begin{align*}
        \dot{x}=\bar{h}(x):=\sum_i \pi(i)h(x,i).
    \end{align*}
    Moreover, suppose the above ODE has a unique globally asymptotically stable equilibrium $x^*$. We have:
    \begin{itemize}
        \item [(a).] If $\sup_t\|x_t\|<\infty$ almost surely, then $x_t\rightarrow x^*$.
        \item [(b).] Suppose $\lim_{c\rightarrow\infty}\frac{\bar{h}(cx)}{c}=h_{\infty}(x)$ exists uniformly on compact sets for some $h_{\infty}\in C(\mbb{R}^n)$. If the ODE $\dot{y}=h_{\infty}(y)$ has origin as the unique globally asymptotically stable equilibrium, then
        \begin{align*}
            \sup_t\|x_t\|<\infty,\ \text{almost surely}.
        \end{align*}
    \end{itemize}
\end{Theorem}

\subsubsection{Kushner-Clark Lemma}
The Kushner-Clark Lemma is a well-known result shown in \cite{kushner2012stochastic}.

Let $\mcal{L}_0$ be an operator projecting a vector onto a compact set $\mcal{X}\subset\mbb{R}^N$. Define an operator $\mcal{L}$ such that for any $x\in\mcal{X}$ and $h:\mcal{X}\rightarrow \mbb{R}^N$ continuous
\begin{align*}
    \mcal{L}(h(x)):=\lim_{\eta\rightarrow 0^+}\frac{\mcal{L}_0(x+\eta h(x))-x}{\eta}.
\end{align*}
Now, consider the following update in $N$ dimensions:
\begin{align*}
    x_{t+1}=\mcal{L}(x_t+\zeta_t(h(x_t)+M_t+\eta_t)).
\end{align*}
Its asymptotic behavior is associated with the following ODE:
\begin{align*}
    \dot{x}=\mcal{L}(h(x)).
\end{align*}

\begin{Theorem}\label{supp2}
Suppose the following assumptions hold:
\begin{itemize}
    \item [(1).] $h(\cdot)$ is a continuous $\mbb{R}^N$-valued function;
    \item [(2).] The sequence $\{\eta_t\}$ is a bounded random sequence with $\lim_t\eta_t=0$ almost surely;
    \item [(3).] The stepsize sequence $\zeta_t$ is such that $\sum_t\zeta_t=\infty$ and $\lim_t\zeta_t=0$;
    \item [(4).] The sequence $\{M_t\}$ satisfies: for any $\epsilon>0$,
    \begin{align*}
        \lim_t \mbb{P}\left(\sup_{n\ge t}\left\|\sum_{t=\tau}^n\zeta_\tau M_\tau\right\|\ge \epsilon\right)=0.
    \end{align*}
\end{itemize}
Then, if the above ODE has a compact set $\mcal{K}^*$ its set of asymptotically stable equilibria, $x_t$ converges almost surely to $\mcal{K}^*$ as $t\rightarrow \infty$.
\end{Theorem}

\subsection{Proof of Theorem \ref{thm:convergen_omega}}
Recall Algorithm~\ref{alg:generic_averagereward} in Appendix~\ref{appendix:algorithms}. For notational convience, in this section, we write $r_t^{(k)}:=f_t^k$ as the $k$-th latent reward for $1\le k\le K$. Let $e_t^{(k)}=f_t^k-\mu_{t-1}^k + Q_{\omega^{t-1}_k}(s_{t+1}, \boldsymbol{\tilde{a}}_{t+1}) - Q_{\omega_k^{t-1}}(s_t, \boldsymbol{a}_t)$, and $y_{t+1}=[e_t^{(1)}\phi_t',\ldots,e_t^{(K)}\phi_t']'\in\mbb{R}^{Km_{\Omega}}$. Here, all quantities with $(k)$ superscript correspond to their $k$-th latent factors. Moreover, since we have assumed $\eta_{\theta,t}=o(\eta_{\omega,t})$, in the spirit of the classical two-time-scale SA analysis in Chapter 6.1 of \citep{borkar2008stochastic}, we can let the policy parameter be fixed as $\theta_t\equiv \theta$ when analyzing the convergence of the critic step, which implies we can show the convergence of $\omega_t$ to some $\omega_{\theta}$ depending on $\theta$. For notational simplicity, we eliminate the notations associated with $\theta$ unless otherwise noted.

The update in Algorithm~\ref{alg:generic_averagereward} in Appendix~\ref{appendix:algorithms} is then expressed as
\begin{align*}
    \mu_{t+1}^{(k)}&=(1-\eta_{\omega,t})\mu_{t}^{(k)}+\eta_{\omega,t}r_{t+1}^{(k)},\\
    \omega_{t+1}^{(k)}&=\omega_t^{(k)}+\eta_{\omega,t} y_{t+1}^{(k)}.
\end{align*}
Here, we will apply Theorem \ref{supp1} as a technical result. To this end, we first verify all the assumptions in Theorem \ref{supp1}. We write the above update as
\begin{align}\label{updateomegaode}
    \mu_{t+1}^{(k)}&=\mu_t^{(k)}+\eta_{\mu,t}\mbb{E}\left(r_{t+1}^{(k)}- \mu_t^{(k)}|\mcal{F}_{t,1}\right)+\eta_{\mu,t}A_{\mu,t+1}^{(k)},\nonumber\\
    \omega_{t+1}^{(k)}&=\omega_t^{(k)}+\eta_{\omega,t}\mbb{E}\left( e_t^{(k)} \phi_t|\mcal{F}_{t,1}\right)+\eta_{\omega,t}A_{\omega,t+1}^{(k)},
\end{align}
where $A_{\mu,t+1}^{(k)}=r_{t+1}^{(k)}-\mbb{E}\left(r_{t+1}^{(k)}|\mcal{F}_{t,1}\right)$ and $A_{\omega,t+1}=y_{t+1}^{(k)}-\mbb{E}\left(e_t^{(k)}\phi_t|\mcal{F}_{t,1}\right)$. Here, $\mbb{E}\left(e_t^{(k)} \phi_t|\mcal{F}_{t,1}\right)$ is Lipschitz continuous in $\omega_t^{(k)}$ and $ \mu_t^{(k)}$. Moreover, by definition, $\{A_{\mu,t}^{(k)}\}$ is a martingale difference sequence satisfying: there exists a constant $C>0$ such that
\begin{align*}
    \mbb{E}\left(\|A_{\mu,t+1}^{(k)}\|^2|\mcal{F}_{t,1}\right)\le C(1+\| \omega_t^{(k)}\|^2+\| \mu_t^{(k)}\|^2),
\end{align*}
given $r_{t+1}^{(k)}$ is uniformly bounded by Assumption \ref{Assump_1}. Furthermore, we will show $\{A_{\omega,t}^{(k)}\}$ is also a martingale difference sequence. Note that by definition,
\begin{align*}
    \mbb{E}\left( y_{t+1}^{(k)}|\mcal{F}_{t,1}\right)=\mbb{E}\left(e_t^{(k)}\phi_t|\mcal{F}_{t,1}\right),
\end{align*}
and
\begin{align}\label{boundAomegasqure}
    \mbb{E}\left(\|A_{\omega,t+1}^{(k)}\|^2|\mcal{F}_{t,1}\right)\le 2\left(\mbb{E}\left(\|y_{t+1}^{(k)}\|^2|\mcal{F}_{t,1}\right)+\|\mbb{E}( e_t^{(k)}\phi_t|\mcal{F}_{t,1})\|^2\right).
\end{align}
Thus, the first term can be bounded over the set $\{\sup_t\|z_t^{(k)}\|\le M\}$ for any $M>0$. Note that
Here, by definition and Jensen's inequality, we have:
\begin{align*}
    \mbb{E}(\|y_{t+1}^{(k)}\|^2|\mcal{F}_{t,1})&=\mbb{E}\left(\|\delta_t^{(k)}\phi_t\|^2|\mcal{F}_{t,1}\right)\\
    &=\mbb{E}\left(\|(r_{t+1}^{(k)}-\mu_t^{(k)}+\phi_{t+1}'\omega_t^{(k)}-\phi_t'\omega_t^{(k)})\phi_t\|^2|\mcal{F}_{t,1}\right)\\
    &\le 3\mbb{E}\left(\left(\|r_{t+1}^{(k)}\phi_t\|^2+\|\mu_t^{(k)}\phi_t\|^2+\|\phi_t(\phi_{t+1}'-\phi_t')\|^2(\omega_t^{(k)})^2\right)|\mcal{F}_{t,1}\right).
\end{align*} 
By our assumption, $\mbb{E}\left(\|\phi_t(\phi_{t+1}'-\phi_t')\|^2|\mcal{F}_{t,1}\right)$ and $\mbb{E}\left(\|\phi_t\|^2|\mcal{F}_{t,1}\right)$ are both uniformly bounded for any $s_t\in\mcal{S},a_t\in\mcal{A}$. Also, due to Assumption \ref{Assump_1}, $\mbb{E}\left(|r_{t+1}^{(k)}|^2|\mcal{F}_{t,1}\right)=\mbb{E}\left(|r_{t+1}^{(k)}|^2|s_t,a_t\right)$ is uniformly bounded. Then, over the set $\{\sup_t\|z_t^{(k)}\|\le M\}$, it holds that: there exists a constant $C>0$,
\begin{align*}
\mbb{E}\left(\|y_{t+1}^{(k)}\|^2|\mcal{F}_{t,1}\right)\cdot\mathds{1}(\sup_t\|z_t^{(k)}\|\le M)\le C.
\end{align*}
Now, the second term in the right-hand side of \eqref{boundAomegasqure} can be directly bounded by $$\|\mbb{E}(e_t^{(k)}\phi_t|\mcal{F}_{t,1})\|^2\le C(1+\| \omega_t^{(k)}\|^2+\| \mu_t^{(k)}\|^2),$$ for some constant $C>0$ since we assume $r_{t+1}^{(k)}$ and $\phi_t$ are bounded by Assumption \ref{Assump_1} and Assumption \ref{Assump_5}. Then, we have: for any $M>0$,
\begin{align*}
    \mbb{E}\left(\|A_{\omega,t+1}^{(k)}\|^2|\mcal{F}_{t,1}\right)\le C(1+\| \omega_t^{(k)}\|^2+\| \mu_t^{(k)}\|^2),
\end{align*}
over the set $\{\sup_t\|z_t^{(k)}\|\le M\}$. This verifies assumption (4) in Theorem \ref{supp1}.

Now, according to Theorem \ref{supp1}, we consider the ODE associated with the asymptotic behavior of \eqref{updateomegaode}:
\begin{align*}
    \dot{ z}^{(k)}=
    \begin{pmatrix}
        \dot{ \mu}^{(k)}\\\dot{ \omega}^{(k)}
    \end{pmatrix}
    =\begin{pmatrix}
        -1&0\\
        -\Phi'D_{\theta}^{s,a}\mbf{1}&\Phi'D_{\theta}^{s,a}(P^{\theta}-I)\Phi
    \end{pmatrix}
    \begin{pmatrix}
         \mu^{(k)}\\ \omega^{(k)}
    \end{pmatrix}
    +
    \begin{pmatrix}
        J^{(k)}(\theta)\\\Phi'D_{\theta}^{s,a}R^{(k)}
    \end{pmatrix}
    ,
\end{align*}
where recall that $D_{\theta}^{s,a}=\text{diag}(d_{\theta}(s)\pi_{\theta}(s,a),s\in\mcal{S},a\in\mcal{A})$ and $J^{(k)}(\theta)$ and $R^{(k)}$ are defined in Theorem \ref{thm:convergen_omega}. In order to apply the conclusions of Theorem \ref{supp1}, we then examine the equilibrium to the above ODE. Denote by $h^{(k)}( z^{(k)})$ the right-hand side of the above ODE. It is straightforward to verify $h^{(k)}(\cdot)$ is Lipschitz, satisfying assumption (1) there. Moreover, by the Perron-Frobenius theorem, the stochastic matrix $P^{\theta}$ has a simple eigenvalue of $1$ while the rest of the eigenvalues have real parts less than $1$. Thus, $P^{\theta}-I$ has a simple eigenvalue of $0$ with all other eigenvalues having negative real parts, so does the matrix $\Phi'D_{\theta}^{s,a}(P^{\theta}-I)\Phi$, given that we assume $\Phi$ is full column rank in Assumption \ref{Assump_5}. The simple eigenvalue of $0$ has eigenvector $\nu$ such that $\Phi\nu=\varrho\mbf{1}$ for some $\varrho>0$ by checking that $\varrho\mbf{1}$ lies in the eigenspace of $D_{\theta}^{s,a}(P^{\theta}-I$ associated with $0$. Consequently, the above ODE is globally asymptotically stable and has its equilibrium satisfying:
\begin{align*}
    \left\{
    \begin{aligned}
    & \mu^{(k)}=J^{(k)}(\theta)\\
    &\Phi'D_{\theta}^{s,a}(R^{(k)}- \mu^{(k)}\mbf{1}+P^{\theta}\Phi\omega^{(k)}-\Phi\omega^{(k)})=0.
    \end{aligned}
    \right.
\end{align*}
Note here that the solution for $ \omega^{(k)}$ has the form $\omega^{(k)}_{\theta}+\varrho\nu$ with any $\varrho>0$ and $\nu\in\mbb{R}^{m_{\Omega}}$ such that $\Phi\nu=\mbf{1}$. Note that by Assumption \ref{Assump_5}, $\Phi\nu\neq \mbf{1}$. This implies $\omega_{\theta}^{(k)}$ is unique such that $\Phi'D_{\theta}^{s,a}(T_{\theta}^{Q,(k)}(\Phi\omega_{\theta}^{(k)})-\Phi\omega_{\theta}^{(k)})=0$ with $T_{\theta}^{Q,(k)}$ given in Theorem \ref{thm:convergen_omega}.

Now, by \citep[Lemma B.1 and B.2]{zhang2018fully}, it implies that $\{\mu_t^{(k)}\}$ and $\{\omega_t^{(k)}\}$ are both bounded almost surely, showing that $\{z_t^{(k)}\}$ is bounded almost surely. Therefore, all assumptions in Theorem \ref{supp1} are satisfied. Consequently, we obtain $\lim_t\mu_t^{(k)}=J^{(k)}(\theta)$ and $\lim_t\omega_t^{(k)}=\omega_{\theta}^{(k)}$ over the set $\{\sup_t\|z_t^{(k)}\|\le M\}$ for any $M>0$, which holds almost surely. Then, it completes the proof for $k$. Since the above arguments hold for any $1\le k\le K$, we obtain the proof for Theorem \ref{thm:convergen_omega}.

\subsection{Proof of Theorem \ref{thm:convergence_theta}}
Let $\mcal{F}_{t,2}=\sigma(\{\theta_\tau\}_{\tau\le t})$ be the $\sigma$-field generated by $\{\theta_\tau\}_{\tau\le t}$. Now, define:
\begin{align*}
    B_{t+1,1}^i&:=A_t^i\psi_t^i-\mbb{E}_{s_t\sim d_{\theta_t},a_t\sim\pi_{\theta_t}}(A_t^i\psi_t^i|\mcal{F}_{t,2}),\\
    B_{t+1,2}^i&:=\mbb{E}_{s_t\sim d_{\theta_t},a_t\sim\pi_{\theta_t}}((A_t^i-A_{t,\theta_t}^i)\psi_t^i)|\mcal{F}_{t,2}),
\end{align*}
where $A_{t,\theta_t}^i$ is defined in \eqref{Apsi} with $\theta=\theta_t$. Recall that the update in Algorithm~\ref{alg:generic_averagereward} can be written with a local projection $\mcal{L}^i$:
\begin{align*}
    \theta_{t+1}^i=\mcal{L}^i(\theta_t^i+\eta_{\theta,t}\mbb{E}_{s_t\sim d_{\theta_t},a_t\sim\pi_{\theta_t}}(A_{t,\theta_t}^i\psi_t^i|\mcal{F}_{t,2})+\eta_{\theta,t}B_{t+1,1}^i+\eta_{\theta,t}B_{t+1,2}^i),
\end{align*}
where $B_{t+1,2}^i=o(1)$ since $\eta_{\theta,t}=o(\eta_{\omega,t})$ by Assumption \ref{Assump_3}, then the critic convergence in Theorem \ref{thm:convergen_omega} is faster thus implying $A_t^i\rightarrow A_{t,\theta}^i$ . Furthermore, let $M_t^i=\sum_{\tau=0}^t\eta_{\theta,\tau}B_{t+1,1}^i$. It is a martingale sequence and since $\{\omega_t^{(k)}\}$, $\{\gamma_{it}(k)\}$, $\{\psi_t^i\}$ and $\{\phi_t\}$ are all bounded, by Assumption \ref{Assump_3}, it holds that
\begin{align*}
    \sum_t\mbb{E}\left(\|M_{t+1}^i-M_t^i\|^2|\mcal{F}_{t,2}\right)=\sum_{t\ge 1}\|\eta_{\theta,t}B_{t+1,1}^i\|^2<\infty,\ \text{almost surely}.
\end{align*}
By the martingale convergence theorem \citep[Proposition VII-2-3(c)]{neveudiscrete}, $\{M_t^i\}$ converges almost surely and it implies that for any $\epsilon>0$,
\begin{align*}
    \lim_t\mbb{P}\left(\sup_{n\ge t}\left\|\sum_{\tau=t}^n\eta_{\theta,t}B_{\tau,1}^i\right\|\ge \epsilon\right)=0
\end{align*}
Now, let 
\begin{align*}
    g^i(\theta_t)=\mbb{E}_{s_t\sim d_{\theta_t},a_t\sim \pi_{\theta_t}}(A_{t,\theta_t}^i\psi_t^i|\mcal{F}_{t,2})=\sum_{s_t\in\mcal{S},a_t\in\mcal{A}}d_{\theta_t}(s_t)\pi_{\theta_t}(s_t,a_t)A_{t,\theta_t}^i\psi_{t,\theta_t}^i.
\end{align*}
In order to apply Theorem \ref{supp2} to show the convergence of $\theta_t^i$, we need to verify that $g^i(\theta_t)$ is continuous in $\theta_t^i$. First, $\psi_t^i$ is continuous in $\theta_t^i$. Next, note that $d_{\theta_t}(s_t)\pi_{\theta_t}(s_t,a_t)$ is the stationary distribution and thus is the solution to the equation $d_{\theta_t}(s)\pi_{\theta_t}(s,a)=\sum_{s'\in\mcal{S},a'\in\mcal{A}}P^{\theta_t}(s',a'|s,a)d_{\theta_t}(s')\pi_{\theta_t}(s',a')$ and $\sum_{s\in\mcal{S},a\in\mcal{A}} d_{\theta_t}(s)\pi_{\theta_t}(s,a)=1$, where $P^{\theta_t}(s',a'|s,a)=P(s'|s,a)\pi_{\theta_t}(s',a')$. The unique solution to this set of linear equations can be verified to be continuous in $\theta_t$ with $\pi_{\theta_t}(s,a)>0$. Moreover, $A_{t,\theta_t}^i$ is continuous in $\theta_t^i$ too as $\omega_{\theta_t}=\sum_{k=1}^K \gamma_{it}(k)\omega_{\theta_t}^{(k)}$, where $\omega_{\theta_t}^{(k)}$ is also continuous due to the fact that it is the unique solution to the linear equation $\Phi'D_{\theta}^{s,a}(T_{\theta}^{Q,(k)}(\Phi\omega_{\theta}^{(k)})-\Phi\omega_{\theta}^{(k)})=0$, which can also be verified to be continuous in $\theta_t$. Therefore, the convergence of $\theta_t^i$ is a direct consequence of applying Theorem \ref{supp2} for each $i\in\mcal{N}$.

\clearpage
\section{Algorithms}\label{appendix:algorithms}

% We list the following practical algorithms with our proposed latent objective discovery.
\begin{itemize}[leftmargin=*]
     
      \item Appendix~\ref{appendix:algorithm_averagerewards}: we include a generic on-policy average-reward actor–critic in Algorithm~\ref{alg:generic_averagereward}. Off-policy variants (e.g., TD3-style) can be obtained by replacing the target action sampling with a target policy and replay buffer.
     
     \item Appendix~\ref{appendix:algorithm_generic}: we include a generic on-policy and off-policy policy gradient algorithm framework in Algorithm~\ref{alg:generic_on} and Algorithm~\ref{alg:generic_off}, respectively.

\end{itemize}

\subsection{Pseudocode of Average-Reward Actor-Critic}\label{appendix:algorithm_averagerewards}

\begin{algorithm}[htbp]
    \centering
	\caption{Actor-Critic Algorithm with Latent Objective Discovery in the Average-Reward Setting}
    \begin{algorithmic}[1] %[1] enables line numbers
		\STATE \textbf{Initialize}: $\{Q_{\omega_k^0}\}_{k=1}^K$, $\{\pi_{\theta_i^0}\}_{i \in \mcal{N}}$, the historical average reward $\{\mu_0^k\}_{k=1}^K$, $\Gamma_0$, $\boldsymbol{f}_0$, the step sizes $\eta_{\omega, t}$ and $\eta_{\theta, t}$. 
		\FOR{$t = 1$ to max-episode-length}
            \FOR{$i \in \mcal{N}$}
                \STATE Observe $s_t$, take action $a_t^i$, receive $r_t^i$
            \ENDFOR
            \STATE Form $\boldsymbol{a}_t$ and $\boldsymbol{r}_t$, observe $s_{t+1}$.
            \STATE {\color{gray} \#  Latent Reward Discovery.}
            % \begin{tcolorbox}[
            %       colback=gray!10,
            %       colframe=gray,
            %       title=Step 1: Latent Reward Discovery,
            %       breakable
            %     ]
            \STATE Online estimation of $\boldsymbol{f}_t$ and $\Gamma_t$ from $\boldsymbol{r}_t$ given $\Gamma_{t-1}$ and $\boldsymbol{f}_{t-1}$ via Eq.~\ref{eq:LOD_estimation}.
            % \end{tcolorbox}
            \STATE {\color{gray} \#  Critic Update.}
            % \begin{tcolorbox}[
            %           colback=orange!10,
            %           colframe=orange!80!black,
            %           title=Step 2: Critic Update,
            %           breakable
            %         ]
            \FOR{$i \in \mcal{N}$}
            \STATE form $\boldsymbol{\tilde{a}}_{t+1}$ by $\tilde{a}^{i}_{t+1} \sim \pi_{\theta_i^{t-1}}(\cdot|s_{t+1})$
            \ENDFOR
            \FOR{$k=1$ to $K$}
                % \STATE 
                \STATE $ \delta_t^k \leftarrow f_t^k-\mu_{t-1}^k + Q_{\omega^{t-1}_k}(s_{t+1}, \boldsymbol{\tilde{a}}_{t+1}) - Q_{\omega_k^{t-1}}(s_t, \boldsymbol{a}_t)$  {\color{gray} \#  Average-Reward Temporal Difference}
                \STATE $\omega_k^t \leftarrow \omega_k^{t-1} + \eta_{\omega, t} \delta_t^k \nabla_{\omega_k}Q_{\omega^{t-1}_k}(s_t, \boldsymbol{a}_t)$
                \STATE $\mu_t^k \leftarrow (1-\eta_{\omega, t}) \mu_{t-1}^k +\eta_{\omega, t} f_t^k$
            \ENDFOR
            % \end{tcolorbox}
            \STATE {\color{gray} \#  Actor Update.}
             % \begin{tcolorbox}[
             %          colback=green!10,
             %          colframe=green!60!black,
             %          title=Step 3: Actor Update,
             %          breakable
             %        ]
             \FOR{$i\in \mcal{N}$}
                % \STATE
                \STATE $Q_i^t \leftarrow \sum_{k=1}^K \gamma^t_{i}(k) Q_{\omega_k^t}(s_t, \boldsymbol{a}_t)$  {\color{gray} \#  Mixed Centralization.}
                \STATE $A_i^t \leftarrow Q_i^t - \sum_{\tilde{a}_t^i}\pi_{\theta_i^{t-1}}(\tilde{a}_t^i|s_t) Q^t_i(s_t, (\tilde{a}_t^i, a_t^{-i}))$
                \STATE $\theta_i^t \leftarrow \theta_i^{t-1} + \eta_{\theta, t} A_i^t \nabla_{\theta_i^{t-1}} \log \pi_{\theta_i^{t-1}}(a_t^i|s_t)$
            \ENDFOR
            % \end{tcolorbox}
		\ENDFOR
	\end{algorithmic}
	\label{alg:generic_averagereward}
\end{algorithm}    

\subsection{Generic On-policy and Off-policy Policy Gradient Algorithms}\label{appendix:algorithm_generic}

\noindent \textbf{On-Policy Multi-agent Policy Gradient.} The algorithm description is provided in Algorithm~\ref{alg:generic_on}.

\begin{algorithm}[htbp]
\centering
\caption{Generic \textbf{On-Policy} Multi-Agent Actor-Critic with Latent Objective Discovery for the Episode Update}
\label{alg:generic_on}
\begin{algorithmic}[1]
\STATE \textbf{Initialize:} stochastic policies $\{\pi_{\theta_i}\}_{i\in\mathcal N}$ and latent value networks $\{V_{\phi_k}\}_{k=1}^K$,  $\Gamma_0$ and $\boldsymbol f_0$.

\STATE Collect $\{s_t,\{o_t^i\}_{i \in \mcal{N}},\boldsymbol a_t,\boldsymbol{r}_t,s_{t+1},\{o_{t+1}^i\}_{i\in \mcal{N}}\}_{t=1}^T$ in the rollout buffer. 
    \STATE Update $\boldsymbol{f}$ and  $\Gamma$ from $\boldsymbol r_{1:T}$. {\color{gray}\# Online Estimation for Latent Objectives}
      \FOR{$k = 1, \ldots, K$}
        \STATE $\hat{A}^k \leftarrow \text{GAE}(V_{\phi_k}, \{f_t^k\}_{t=1}^T)$ {\color{gray} \#  Mixed Centralized Advantage Function via GAE.}
    \ENDFOR
    \STATE {\color{gray} \#  Actor update given the current $\Gamma_T$.}
    \STATE Update each policy with $\pi_{\theta_i}$ by maximizing clipped PPO or TRPO objective based on $\sum_{k=1}^K \gamma_{i}(k) \hat{A}^k$
    \STATE {\color{gray} \#  Critic update.}
    \FOR{$k = 1, \ldots, K$}
    \STATE Evaluate GAE return: $\hat{R}_t^k = \hat{A}^k(s_t, \boldsymbol{a}_t) + V_{\phi^{\text{old}}_k}(s_t)$
    \STATE Update value function with $V_{\phi_k}$ by minimizing $\frac{1}{T} \sum_{t=1}^T \left[(V_{\phi_k}(s_t)-\hat{R}^k_t)^2\right]$
    \ENDFOR
% \STATE $\Gamma_0 \leftarrow \Gamma_T$ and $\boldsymbol{f}_0 \leftarrow \boldsymbol{f}_T$  {\color{gray} \#  Episode-level warm start.}
\end{algorithmic}
\end{algorithm}

\noindent \textbf{Off-Policy Multi-agent Policy Gradient.}  We define $\mu_{\theta_i}$ as the deterministic policy. The latent critic loss with the TD target under the latent objective discovery are written as
\begin{align}
     \mathcal{L}_Q(\omega)=\frac{1}{B} \sum_{t \in \mathcal{B}} \sum_{k=1}^K\left(Q_{\phi_k}\left(s_t, a_t\right)-y_t^k\right)^2, \quad y_t^k=f_t^k+\gamma Q_{\bar{\phi}_k}\left(s_{t+1}, \boldsymbol{a}^{-}_{t+1}\right),
\end{align}
where $Q_{\bar{\phi}_k}$ is the target network, which determines $\boldsymbol{a}^{-}_{t+1}$ as a greedy action, and is updated periodically.  The algorithm description is provided in Algorithm~\ref{alg:generic_off}.

\begin{algorithm}[htbp]
\centering
\caption{Generic \textbf{Off-Policy} Multi-Agent Actor-Critic with Latent Objective Discovery for the Episode Update}
\label{alg:generic_off}
\begin{algorithmic}[1]
\STATE \textbf{Initialize:} deterministic actors $\{\mu_{\theta_i}\}_{i\in\mathcal N}$ and latent centralized critics $\{Q_{\phi_k}\}_{k=1}^K$ with targets,  $\Gamma_0$ and $\boldsymbol f_0$.

\FOR{$t = 1, \ldots, T$}
    \STATE Sample $a_t^i=\mu_{\theta_i}(o_t^i)+\mathcal N_t^i$, execute $\boldsymbol a_t$, receive $\boldsymbol r_t$, observe $s_{t+1}, \{o_{t+1}^i\}_{i \in \mcal{N}}$; store transitions in replay buffer.
 \STATE {\color{gray}\# Update $\Gamma_t$ in Latent Objective Discovery}
    \STATE Update $\Gamma_t$ and $\boldsymbol{f}_t$ from $\boldsymbol r_t$ given $\Gamma_{t-1}$ and $\boldsymbol{f}_{t-1}$.
    \STATE Sample $\{s_b,\{o_b^i\}_{i\in \mcal{N}},\boldsymbol a_b,\boldsymbol r_b,s_{b+1},\{o_{b+1}^i\}_{i \in \mcal{N}}\}_{b=1}^B$.
    \STATE Decompose $\boldsymbol f_b$ from $\boldsymbol r_b$ via $\Gamma_t$. {\color{gray}\# Projection}
    \STATE {\color{gray}\# Critic update via a general TD target $Q_k$}
    \STATE Evaluate TD target $Q_k$ (DDPG/TD3) and update $Q_{\phi_k}$
    % \vspace{-3mm}
    $$
    \min_{\phi_k} \frac{1}{B}\sum_b (f_b^k + \gamma\,Q_{\bar{\phi}_k}(s_{b+1},\boldsymbol{a}_{b+1}) - Q_{\phi_k}(s_b,\boldsymbol a_b))^2,
   $$
   where $Q_{\bar{\phi}_k}$ is the target value network.
   % \vspace{-4mm}
    \STATE {\color{gray}\# Actor update given the current $\Gamma_t$}
    \STATE Evaluate mixed centralized critics for each agent $i$:
    \vspace{-3mm}
    $$
    \bar Q_i(s,\boldsymbol a)=\sum_{k=1}^K \gamma_{i,t}(k)\,Q_{\phi_k}(s,\boldsymbol a).
    $$
    \vspace{-4mm}
    \STATE Update each policy $\mu_{\theta_i}$ via the gradient:
    \vspace{-3mm}
    $$
    \nabla_{\theta_i}J(\theta_i)=
    \frac{1}{B}\sum_{b=1}^B
    \nabla_{\theta_i}\mu_{\theta_i}(o_b^i)\,
    \nabla_{a^i}\bar Q_i(s_b,\boldsymbol a_b)\big|_{a^i \sim \mu_{\theta_i}}.
    $$
    \vspace{-4mm}
    \STATE Update target actors and value networks by Polyak averaging.
\ENDFOR
\end{algorithmic}
\end{algorithm}

\clearpage
\section{Experiment Details}\label{appendix:experiment}

\subsection{Multi-Agent Particle Environment~(MPE)}\label{appendix:experiment_MPE}

\paragraph{Environment Description.} The Multi-Agent Particle Environments~(MPE) comprise a collection of two-dimensional navigation tasks. We consider two coordination-focused scenarios:\texttt{simple\_spread} and \texttt{Speaker\_listerner}. Agents receive observations represented by high-level feature vectors, including the relative positions of other agents and landmarks, and select actions to navigate in two dimensions. All scenarios are fully observable except \texttt{Speaker\_listerner}, which additionally involves binary communication. Because rewards depend strongly on joint actions, MPE provides a benchmark for evaluating multi-agent coordination and learning under non-stationarity. Ineffective coordination among agents can substantially degrade performance.

\begin{itemize}[leftmargin=*]
    \item  \texttt{simple\_spread:} In this task, $N$ agents learn to cover $N$ landmarks while avoiding collisions. Each agent observes its own position and velocity, together with the relative positions of the other agents and all landmarks, and selects from five discrete movement actions. The reward 
    combines a shared coverage component, the negative sum of the distances from each landmark to its nearest agent, with agent-specific collision penalties. Therefore, the rewards are heterogenous.

    \item \texttt{Speaker\_listerner:} In this task, a static speaker must communicate a target landmark to a listener in an environment containing three landmarks. Both agents receive a shared reward equal to the negative Euclidean distance between the listener and the target. The speaker observes only the target’s colour, whereas the listener observes its own velocity, the relative positions of all landmarks, and the speaker’s message. The speaker learns to encode the target using three discrete communication actions, while the listener selects from the five standard discrete movement actions used in MPE.
\end{itemize}

\paragraph{Implementation Details.} Our implementations are directly adapted from HARL framework~\citep{zhong2024heterogeneous} and deploy the same default hyperparameters in both MAPPO and LOD-MAPPO. For the LOD module, we choose $\tau=0.2$ for the soft updating of $\Gamma$ after the line search as the ablation study and set $K=2$ ($K\leq N$) due to the underlying objectives in the MPE environment. We choose window size as 3, indicating that we estimate $\Gamma$ and $f_t$ by using data across the last $T \times 3$ steps.

\begin{figure}[htbp]
    \centering
    \begin{subfigure}
        \centering
        \includegraphics[width=0.32\textwidth]{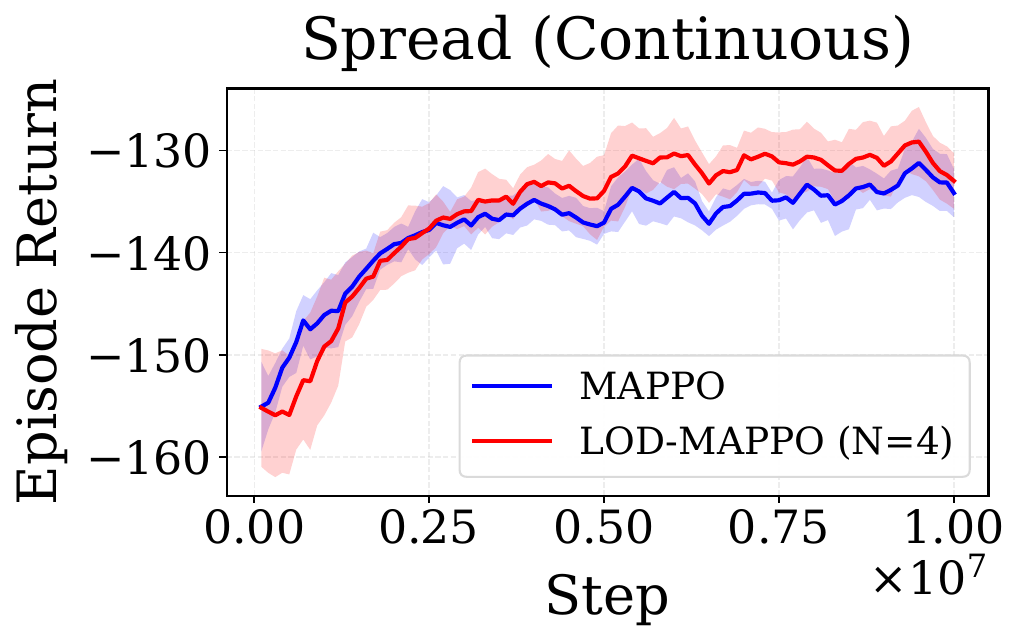}
    \end{subfigure}
    \hfill
    \begin{subfigure}
        \centering
        \includegraphics[width=0.32\textwidth]{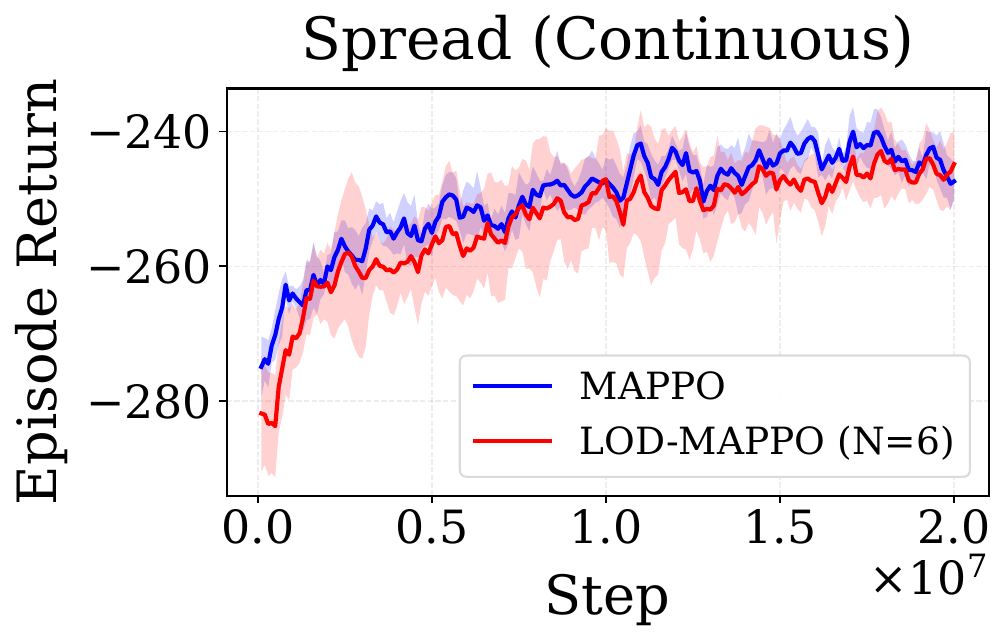}
    \end{subfigure}
    \begin{subfigure}
        \centering
        \includegraphics[width=0.32\textwidth]{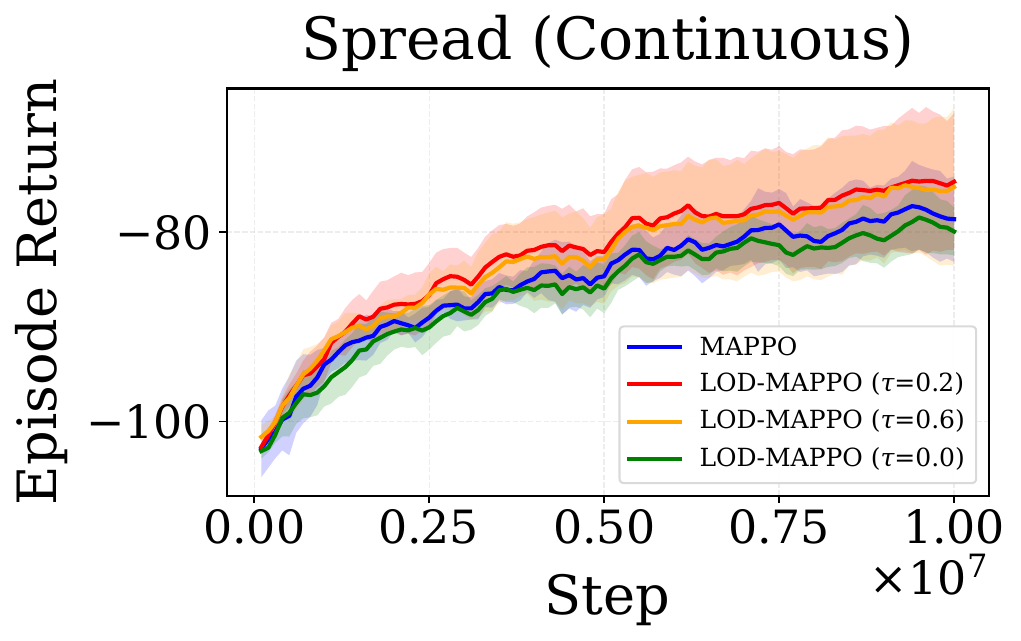}
    \end{subfigure}
    \hfill
    \vspace{-2mm}
    \caption{Ablation study in Spread (Continuous) across $N$ and $\tau$ in the soft updating of $\Gamma$.}
    \label{fig:MPE_ablation}
\end{figure}

\paragraph{Ablation Study.} In Figure~\ref{fig:MPE_ablation}, we focus on spread~(continuous) by increasing $N$ and varying $\tau$, the hyperparameter that controls the soft updating of $\Gamma$ via $\Gamma_{t+1} = (1-\tau)\Gamma_t + \tau \Gamma$. With four agents, LOD-MAPPO generally achieves higher mean returns during the middle and later stages of training, while with six agents, LOD-MAPPO performs similarly to MAPPO. Setting $\tau=0$ freezes the randomly initialized mixing weights, while increasing $\tau$ renders more emphasis on the newly collected data in estimating $\Gamma_{t+1}$. Figure~\ref{fig:MPE_ablation} shows that an overly large or small $\tau$, e.g., $\tau=0.0$ or 0.6 are inferior to $\tau=0.2$ in MPE tasks. After tuning the hyperparameters, we fix them across the remaining MPE tasks.

\clearpage
\subsection{Level-Based Foraging~(LBF) and Multi-Robot Warehouse~(RWARE)}\label{appendix:experiment_LBFRWARE}

\paragraph{Environment Descriptions.} We choose Level-Based Foraging~(LBF) and Multi-Robot Warehouse~(RWARE) to test our LOD-MAPPO algorithms, where the agents have access to heterogeneous rewards.

\begin{itemize}[leftmargin=*]
    \item  \texttt{LBF:} The Level-Based Foraging environment consists of tasks focusing on the coordination of involved agents. The task for each agent is to navigate the grid-world map and collect items. Each agent and item is assigned a level and items are randomly scattered in the environment. In order to collect an item, agents have to choose a certain action next to the item. However, such collection is only successful if the sum of involved agents’ levels is equal or greater than the item level. Agents receive rewards equal to the level of the collected item. The tasks that were selected are denoted first by the grid-world size, $15 \times 15$ means a 15 by 15 grid-world. Rewards are heterogeneous in this environment. The \textbf{cooperation version } of LBF requires agents to coordinate to be able to gather any reward, making the task very hard to solve given the sparsity of the rewards.

    \item \texttt{RWARE:} The multi-robot warehouse environment is a set of collaborative, partially observable multi-agent tasks simulating a warehouse operated by robots. Each agent controls a single robot aiming to collect requested shelves. At all times, $N$ shelves are requested and each timestep a request is delivered to the goal location, a new (currently unrequested) shelf is uniformly sampled and added to the list of requests. Agents are only rewarded whenever an agent is delivering a requested shelf to a goal position. Therefore, a very specific and long sequence of actions is required to receive any non-zero rewards, making this environment \textbf{very sparsely rewarded}.
\end{itemize}

\paragraph{Implementation Details.} Our implementations are directly adapted from EpyMARL framework~\citep{papoudakis2021Benchmarking} and deploy the same default hyperparameters in both MAPPO and LOD-MAPPO. In LBF, we choose $\tau=0.2$, while in RWARE-tiny, we set $\tau=1\times10^{-3}$. By contrast, we find a smaller $\tau$ in RWARE-small can achieve better performance after a line search. This  is because RWARE-small is even harder than RWARE-tiny, which requires a more stable updating (with a smaller $\tau$) for $\Gamma_t$. The window size is 1.

\begin{figure}[htbp]
    \centering
    \begin{subfigure}
        \centering
        \includegraphics[width=0.22\textwidth]{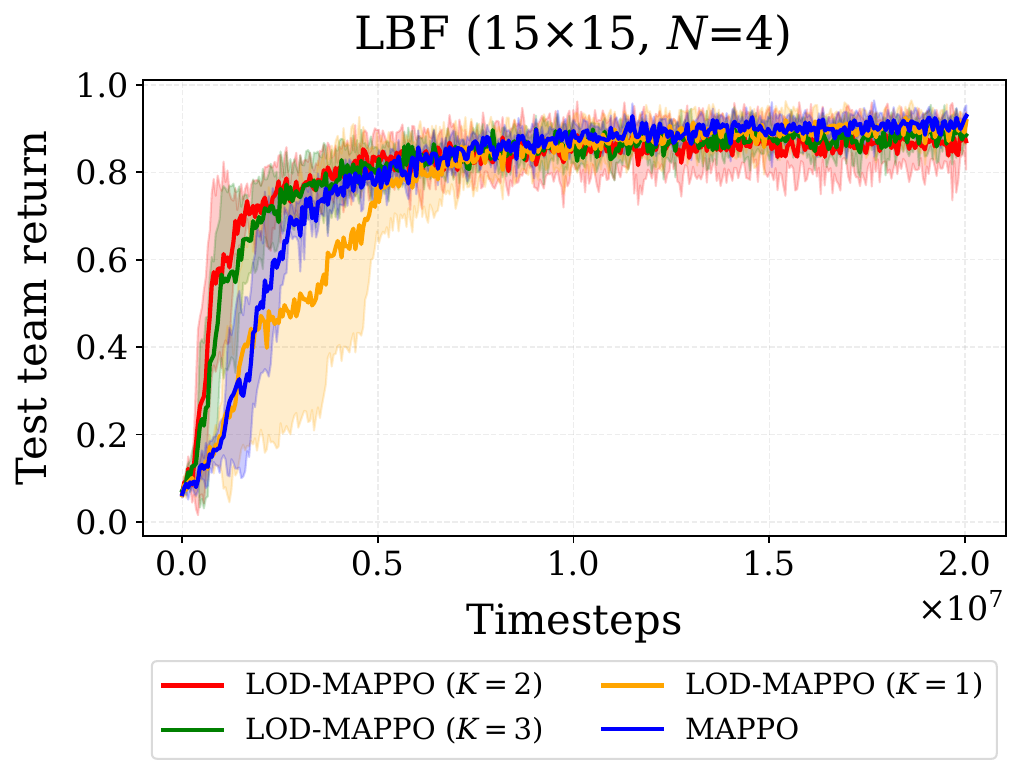}
    \end{subfigure}
    \hfill
    \begin{subfigure}
        \centering
        \includegraphics[width=0.22\textwidth]{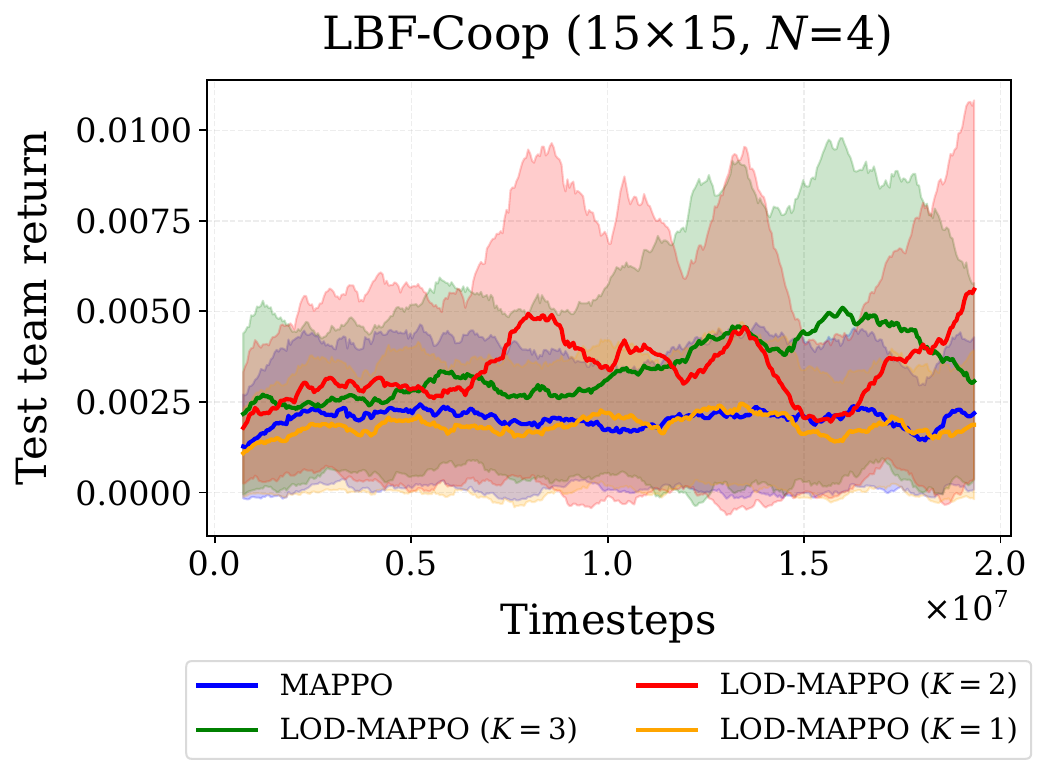}
    \end{subfigure}
    \begin{subfigure}
        \centering
        \includegraphics[width=0.22\textwidth]{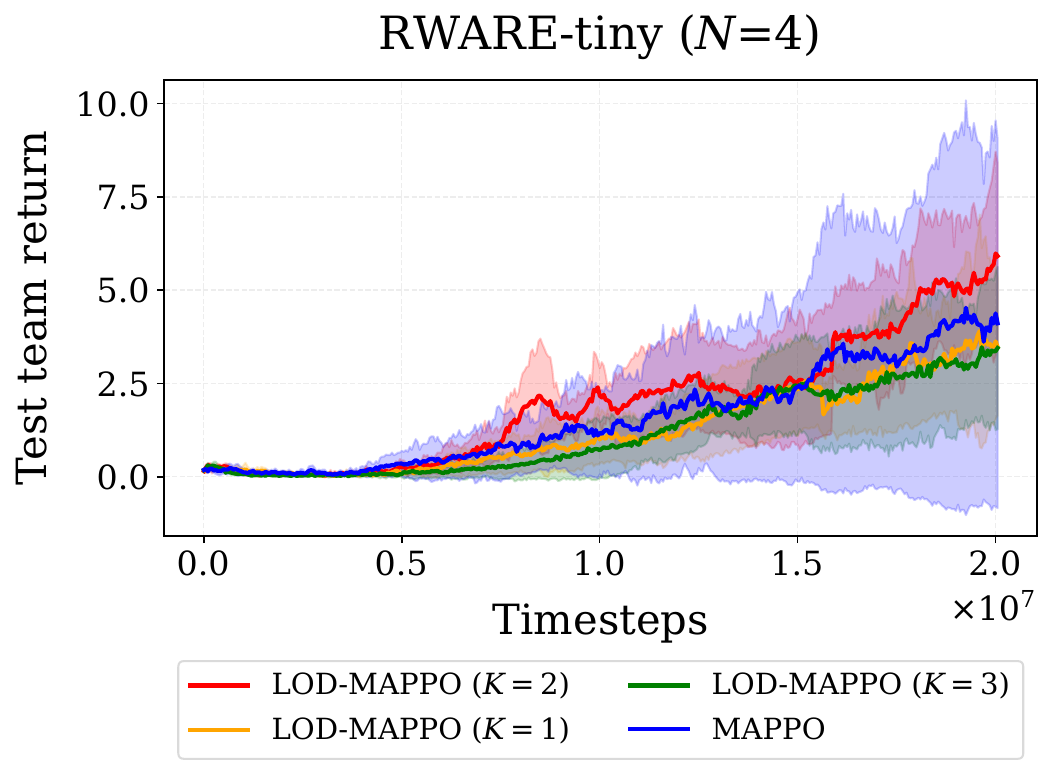}
    \end{subfigure}
    \hfill
    \begin{subfigure}
        \centering
        \includegraphics[width=0.22\textwidth]{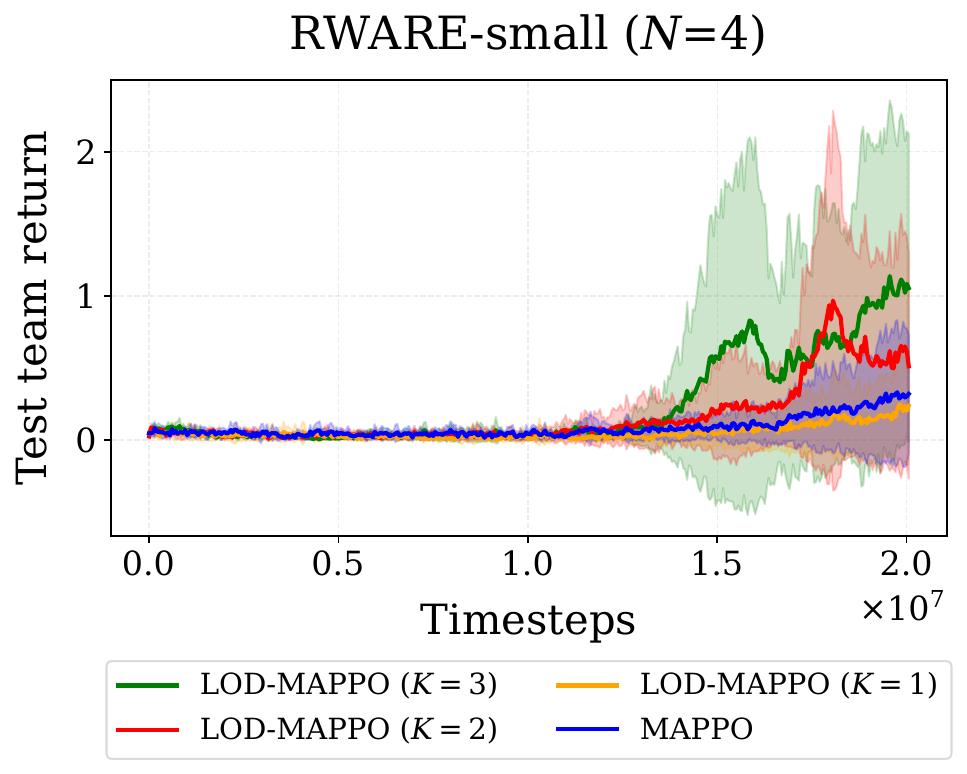}
    \end{subfigure}
    \vspace{-5mm}
    \caption{Comparison between LOD-MAPPO across a wide range of $K$ and MAPPO in LBF and RWARE environments. The learning curves are the means averaged across five random seeds, and the shades indicate standard deviations.}
    \label{fig:LBF_RWARE_shade}
\end{figure}

\paragraph{Learning curves with shades.} Figure~\ref{fig:LBF_RWARE_shade} suggests the learning curves of LOD-MAPPO and MAPPO with the shades that indicate the standard deviations. It showcases that algorithms in LBF-Coop have more variations than LBF, indicating a more difficult environment. In addition, algorithms in RWARE also have very large variations due to the very sparse rewards in this environment.   

\clearpage
\subsection{Agentic RL}\label{appendix:experiment_agenticRL}
\subsubsection{Agent Lightning ChartQA}

We implement the ChartQA experiment based on the Agent Lightning
agentic RL framework~\citep{he2026agent}. The workflow consists of five
successive agent nodes corresponding to the original ChartQA pipeline:
\emph{Chart Analyzer}, \emph{Data Extractor}, \emph{Answer Calculator},
\emph{Answer Checker}, and \emph{Answer Refiner}. Each agent selects one
strategy from its role-specific strategy set, and the sequence of
decisions determines the final answer. We treat these five workflow
nodes as cooperating heterogeneous agents and apply MAPPO and
LOD-MAPPO for policy optimization.

\paragraph{Language Model and Environment.}
We use Qwen2.5 as the underlying language model for all
agents. The model is used only to provide LLM-based
grounding of strategy effectiveness. Specifically, for each role and
strategy, we obtain empirical strategy quality estimates by evaluating
the LLM on ChartQA tasks. The ChartQA environment contains 200
generated tasks with diverse question types, including maximum,
minimum, summation, difference, average, and comparison queries.

\begin{tcolorbox}[
    colback=gray!5,
    colframe=gray!40,
    boxrule=0.5pt,
    arc=3pt,
    title={Example of LLM-based Strategy Grounding},
    fonttitle=\bfseries,
    width=\linewidth
]

\textbf{User Prompt}

\vspace{2pt}

\begin{tcolorbox}[
    colback=white,
    colframe=blue!30,
    boxrule=0.4pt,
    arc=2pt
]
\textbf{Role:} Data Extractor

\textbf{Strategy:} Extract only queried values.

\vspace{4pt}

\textbf{Chart:}

C1: 35, \quad C2: 82, \quad C3: 57, \quad C4: 41

\vspace{4pt}

\textbf{Question:}

What is the average value of C1, C2, and C3?

\vspace{4pt}

Return the chart values needed to answer the question.
\end{tcolorbox}

\vspace{6pt}

\textbf{Qwen2.5 Output}

\vspace{2pt}

\begin{tcolorbox}[
    colback=white,
    colframe=green!40,
    boxrule=0.4pt,
    arc=2pt
]
The queried values are:

\[
\mathrm{C1}=35,\qquad
\mathrm{C2}=82,\qquad
\mathrm{C3}=57.
\]
\end{tcolorbox}

\vspace{6pt}

\textbf{Evaluation}

The extracted values are compared with the ground-truth chart values.
The role--strategy pair receives a binary success score of $1$ if the
extracted information is correct and $0$ otherwise. Aggregating these
scores over ChartQA tasks produces the empirical strategy effectiveness
used for RL training.

\end{tcolorbox}

\paragraph{RL Training Configuration.}
All methods are trained using MAPPO or LOD-MAPPO with the same hyperparameters. For MAPPO, we use
a batch size of 128, four MAPPO optimization epochs, and a minibatch size
of 256. The actor and critic learning rates are set to $10^{-3}$ and
$3\times10^{-3}$, respectively. We use PPO clipping coefficient
$\epsilon=0.2$, entropy coefficient $0.003$, value loss coefficient
$0.5$, and gradient clipping with maximum norm $0.5$. The hidden
dimension of all policy and value networks is 64. For LOD-MAPPO, we evaluate different numbers of latent objectives with
$K\in\{1,2,3\}$. The latent loading matrix is estimated from
counterfactual local advantages according to Eq.~\ref{eq:local_advantage} while using ridge-regularized
factorization. We use ridge coefficient $10^{-3}$, update the loading
matrix every 256 environment steps using a window of 1000 recent
samples, and optimize the factorization for 400 iterations with learning
rate $0.08$.

\paragraph{Reward and Evaluation.}
All agents receive the same binary task-level reward from the ChartQA
workflow. The reward is one when the final answer produced by the
complete workflow is correct and zero otherwise. For LOD-MAPPO, this
shared reward is first converted into agent-specific counterfactual
local advantages, which are then factorized into latent objective
signals for policy optimization. We train each method with five random seeds and report the
moving-average success rate during training. The curves are smoothed
using a moving window of 300 training steps, and the shaded regions
represent the standard error across seeds.

\subsubsection{Agent Lightning MuSiQue}

We also implement the MuSiQue experiment based on the Agent Lightning
agentic RL framework~\citep{he2026agent}. The workflow consists of four
successive retrieval decision positions, which we treat as cooperating
agents. At each step, the corresponding agent selects one retrieval
strategy from four alternatives: \emph{Direct}, which searches for the
main fact needed to answer the question; \emph{Decompose}, which targets
an unresolved intermediate relation; \emph{Entity}, which focuses on a
relevant entity and relation; and \emph{Evidence}, which searches for
information missing from the currently available evidence. The selected
strategy is converted into a search query and used to retrieve supporting
information. We treat the four retrieval turns as heterogeneous agents
and apply MAPPO and LOD-MAPPO for policy optimization.

\begin{tcolorbox}[
    colback=gray!5,
    colframe=gray!40,
    boxrule=0.5pt,
    arc=3pt,
    title={Example of LLM-based Strategy Grounding},
    fonttitle=\bfseries,
    width=\linewidth
]

\textbf{User Prompt}

\vspace{2pt}

\begin{tcolorbox}[
    colback=white,
    colframe=blue!30,
    boxrule=0.4pt,
    arc=2pt
]

\textbf{Question:}

Which city was the author of \emph{Book A} born in?

\vspace{4pt}

\textbf{Evidence Already Found:}

The author of \emph{Book A} is John Smith.

\vspace{4pt}

\textbf{Turn:}

2 of 4

\vspace{4pt}

\textbf{Search Strategy:}

ENTITY: center the query on the relevant named entity and the relation
needed next.

\vspace{4pt}

Generate the next search query. Output only the query.

\end{tcolorbox}

\vspace{6pt}

\textbf{Qwen2.5 Output}

\vspace{2pt}

\begin{tcolorbox}[
    colback=white,
    colframe=green!40,
    boxrule=0.4pt,
    arc=2pt
]
John Smith birthplace city
\end{tcolorbox}

\vspace{6pt}

\textbf{Evaluation}

The generated query is used for retrieval over candidate passages.
The strategy receives success score $1$ if the retrieved top-$k$
passages contain an unresolved supporting paragraph, and $0$ otherwise.
Aggregating these scores over MuSiQue tasks produces empirical strategy
effectiveness estimates.
\end{tcolorbox}

\paragraph{Language Model and Environment.}
We also use Qwen2.5 as the underlying language model for all retrieval agents.
% The model is used only for LLM-based grounding of strategy effectiveness.
Specifically, for each retrieval turn and strategy, Qwen generates a
search query conditioned on the current question and available evidence.
The generated query is evaluated using the MuSiQue supporting-paragraph
annotations: a strategy succeeds if its retrieved passages contain an
unresolved supporting paragraph. Aggregating these outcomes over
benchmark tasks provides empirical strategy effectiveness estimates for
RL training.

The MuSiQue environment contains 200 real benchmark tasks with
multi-hop questions requiring information aggregation from multiple
sources. Each task contains supporting evidence annotations and is
categorized according to its hop number. The retrieval process consists
of four temporal decision steps, where each step selects among the same
four retrieval strategies.

\paragraph{RL Training Configuration.} The hyperparameters of MAPPO and LOD-MAPPO are set the same as ChartQA above.

\paragraph{Reward and Evaluation.} The reward and evaluation strategy are also same as ChartQA. We train each method with five random seeds and report the moving-average
success rate during training. The curves are smoothed using a moving
window of 300 training steps, and shaded regions represent the standard
error across seeds.
\end{document}